\documentclass[conference]{IEEEtran}
\usepackage{times}
\usepackage{graphicx}
\usepackage{wrapfig}
\usepackage{amsmath}
\usepackage{lipsum}
\usepackage[numbers]{natbib}
\usepackage{multicol}
\usepackage{booktabs} 
\usepackage{array} 
\usepackage{tabularx} 
\usepackage{ifthen}
\usepackage{xcolor}
\let\labelindent\relax
\usepackage{enumitem}
\usepackage{xspace}
\usepackage{float}
\usepackage{dblfloatfix} 
\usepackage[bookmarks=true]{hyperref}
\usepackage{soul} 
\usepackage{xcolor} 

\sethlcolor{yellow} 

\newcommand{\MethodName}{\text{\textbf{RialTo}}\xspace}

\newboolean{draft-mode}
\newcommand{\anonymous}{0}
\setboolean{draft-mode}{false}

\DeclareRobustCommand{\asnote}[1]{\ifthenelse{\boolean{draft-mode}}{\textcolor{cyan}{\textbf{Anthony:} #1}}{}}
\DeclareRobustCommand{\pulkit}[1]{\ifthenelse{\boolean{draft-mode}}{\textcolor{blue}{\textbf{Pulkit:} #1}}{}}
\DeclareRobustCommand{\marcel}[1]{\ifthenelse{\boolean{draft-mode}}{\textcolor{red}{\textbf{Marcel:} #1}}{}}
\DeclareRobustCommand{\abhishek}[1]{\ifthenelse{\boolean{draft-mode}}{\textcolor{orange}{\textbf{Abhishek:} #1}}{}}
\begin{document}

\title{Reconciling Reality through Simulation: A Real-to-Sim-to-Real Approach for Robust Manipulation}

\ifthenelse{\anonymous > 0}{
\author{Author Names Omitted for Anonymous Review. Paper-ID [325]}

}{
\author{%
\textbf{Marcel Torne}$^{1}$ \quad \textbf{Anthony Simeonov}$^{1,4}$ \quad \textbf{Zechu Li}$^{1,3,4}$ \quad \textbf{April Chan}$^{1,4}$ \\
\textbf{Tao Chen$^{1,4}$ } \quad \textbf{Abhishek Gupta}$^{2*}$ \quad \textbf{Pulkit Agrawal}$^{1,4*}$\\
$^1$Massachusets Institute of Technology \quad $^2$University of Washington \quad $^3$TU Darmstadt \\ $^4$ Improbable AI Lab 
}
}

\maketitle

\begin{abstract}
Imitation learning methods need significant human supervision to learn policies robust to changes in object poses, physical disturbances, and visual distractors. Reinforcement learning, on the other hand, can explore the environment autonomously to learn robust behaviors but may require impractical amounts of unsafe real-world data collection. To learn performant, robust policies without the burden of unsafe real-world data collection or extensive human supervision, we propose \MethodName, a system for robustifying real-world imitation learning policies via reinforcement learning in ``digital twin" simulation environments constructed on the fly from small amounts of real-world data. To enable this real-to-sim-to-real pipeline, \MethodName proposes an easy-to-use interface for quickly scanning and constructing digital twins of real-world environments. We also introduce a novel ``inverse distillation" procedure for bringing real-world demonstrations into simulated environments for efficient fine-tuning, with minimal human intervention and engineering required. We evaluate \MethodName across a variety of robotic manipulation problems in the real world, such as robustly stacking dishes on a rack, placing books on a shelf, and six other tasks. \MethodName increases (over 67\%) in policy robustness without requiring extensive human data collection. Project website and code at 
\ifthenelse{\anonymous > 0}{
REDACTED.
}{
 \url{https://real-to-sim-to-real.github.io/RialTo/}.
}
\end{abstract}

\let\thefootnote\relax\footnotetext{* Equal advising}

\begin{figure*}[h!tbp]
    \centering
    \includegraphics[width=0.95\linewidth]{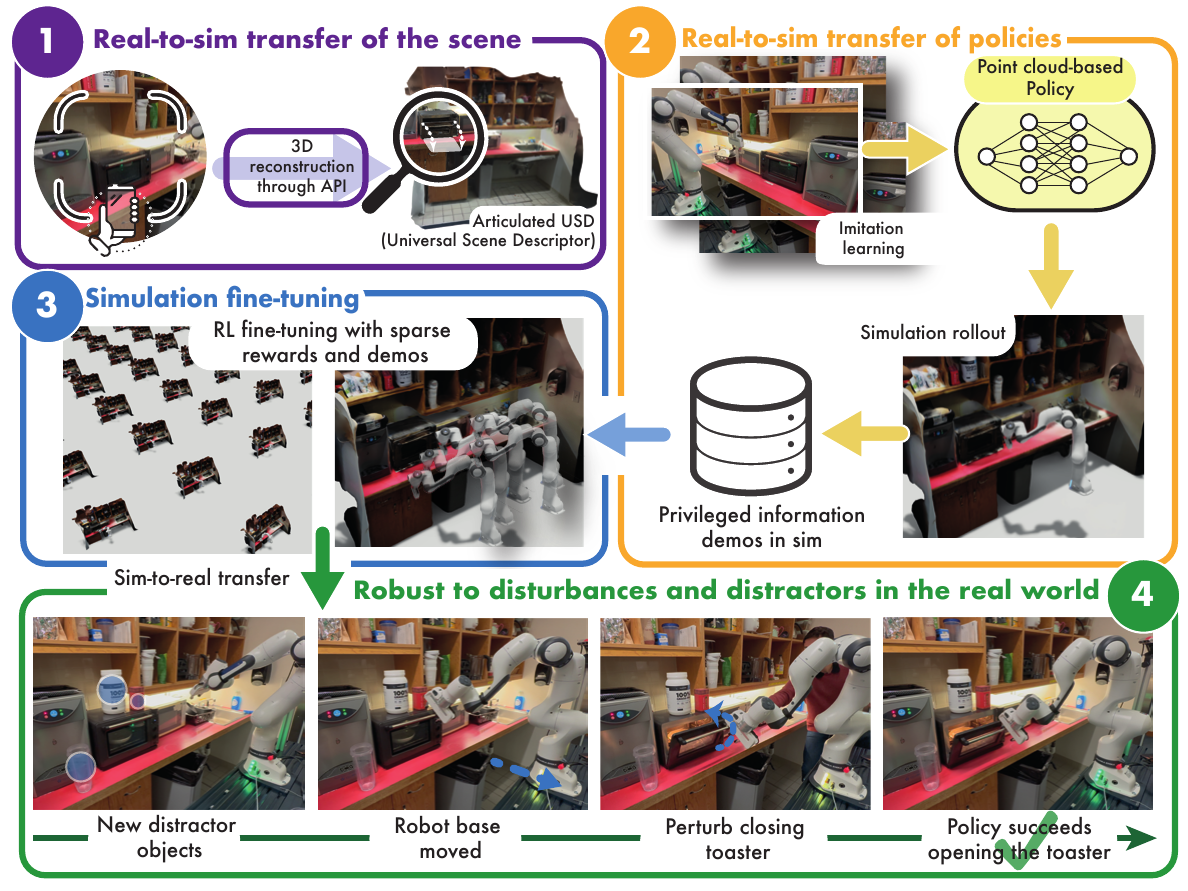}
    \caption{\footnotesize{\textbf{\MethodName system overview}. 1) Transfer the real-world scene to the simulator through an easy-to-use API (see Section \ref{sec:r2simscenes}). 2) Transfer a policy learned from real-world demonstrations to collect a set of demonstrations with privileged information in simulation. We note this step is optional, and \MethodName is compatible with skipping this step and providing demonstrations in simulation (see Section \ref{subsec:r2simpolicies}) 3) Use the collected set of demonstrations to bias exploration in the RL fine-tuning with sparse rewards of a state-based policy (see Section \ref{sec:robustification}) 4) Perform teacher-student distillation and deploy the policy in the real world obtaining robust behaviors (see Section \ref{sec:teacherstudentcotraining}). }} 
    \label{fig:teaser}
\end{figure*}

\section{Introduction}
\label{sec:introduction}
Imagine a robot that can de-clutter kitchens by putting dishes on a dish rack. Consider all the environmental variations that might be encountered: different configurations of plates or changes in rack positions, a plate unexpectedly slipping in the gripper during transit, and visual distractions, including clutter and lighting changes. For the robot to be effective, it must robustly solve the task across the various scene and object perturbations, without being brittle to transient scene disturbances. Our desiderata is a framework that makes it \textit{easy} for humans to program the robot to achieve a task \textit{robustly} under these variations or disturbances. To be a scalable choice for deployment, the framework should not make task-specific assumptions and must seamlessly apply to many tasks.

To design these types of robust robot controllers, one could attempt to train policies across a massive range of scenes and with highly variable objects~\cite{chen2023visual, gong2023arnold}. This is hard-pressed to provide a scalable solution to robotic learning for two reasons - (1) it is challenging to actually collect or synthesize data across a massive range of scenes since content creation can be challenging in simulation and data collection can be challenging for the real world, (2) a widely general, robust policy may be overly conservative, lowering its performance on the specific target domains encountered on deployment. Alternatively, we suggest that to maximally benefit a \emph{specific} user, it is more critical that the robot achieves high success in their \emph{particular} home environment, showing robustness to various local disturbances and distractors that might be encountered in this setting. With this in mind, our goal is to develop a robot learning technique that requires minimal human effort to synthesize visuomotor manipulation controllers that are extremely robust for task performance in deployment environments. The question becomes - how do we acquire these robust controllers without requiring prohibitive amounts of effort for data collection or simulation engineering?

A potential technique for data-driven learning of robotic control policies is to adopt the paradigm of imitation learning (IL), learning from expert demonstration data~\cite{schaal2003computational, ratliff2007imitation, hussein2017imitation}. However, controllers learned via imitation learning tend to exhibit limited robustness unless a large number of demonstrations are collected. Furthermore, imitation learning does not learn to recover from mistakes or out-of-distribution disturbances unless such behaviors were intentionally demonstrated. This makes direct imitation learning algorithms unsuitable for widespread, robust deployment in real-world scenarios. 

The alternative paradigm of reinforcement learning (RL) allows robots to train on self-collected data, reducing the burden on humans for \emph{extensive} data collection~\cite{kober2013reinforcement} and to discover robust recovery behaviors \emph{beyond} a set of pre-collected demonstrations (e.g., re-grasping when an object is dropped, re-aligning when an object moves in the gripper, adjusting to external perturbations, etc. --- see examples in Fig.~\ref{fig:teaser}). However, directly performing RL in the real world is prohibitively slow, often results in unsafe data collection, and is challenging due to problems like resets and reward specification~\cite{zhu2020ingredients}. Therefore, currently, it's impractical in many cases to employ RL for learning robust control policies directly in the real world. Simulation, on the other hand, offers the ability to collect significant amounts of data broadly, cheaply, safely, and with privileged information ~\cite{chen2023visual, lee2020learning, song2023reaching, margolis2022rapid, andrychowicz2020learning}. However, manually constructing geometrically, visually, and physically realistic simulation environments for problems like robotic manipulation in the home can be time and labor-intensive, making it an impractical alternative at scale.

To safely and efficiently learn robust manipulation behaviors, our key insight is to train RL controllers on \emph{quickly} constructed simulation scenes. By leveraging a video from the target deployment domain, we can obtain scenes complete with accurate geometry and articulation that reflect the appearance and kinematics of the real world.
These ``in-domain" simulation environments can serve as a sandbox to safely and quickly learn robust policies across various disturbances and distractors, without requiring expensive exploration in the real world. We show how imitation learning policies trained with small numbers of real-world demonstrations can be robustified via large-scale RL fine-tuning in simulation on these constructed simulation environments, using minimal amounts of human effort in terms of environment design and reward engineering. To remove the burden of reward engineering, we leverage a set of real-world demonstrations that bootstrap efficient fine-tuning with reinforcement learning. These real-world demonstrations help narrow the sim-to-real gap and increase the performance of our policies, as shown in Section \ref{subsec:cotrainingresults}. However, transferring real-world demonstrations into simulation is non-trivial because we do not have access to the Lagrangian state of the environment (e.g. object poses). We therefore propose a new ``inverse-distillation" technique that enables transferring real-world demos into the simulation.
After using RL in constructed simulation environments to robustify the real-world imitation learning policies, the fine-tuned policies can be transferred back to the real world with significantly improved success rates and robustness to test-time disturbances. 

Overall, our pipeline simultaneously improves the effectiveness of both reinforcement \emph{and} imitation learning. RL in simulation helps make imitation learning policies deployment-ready without requiring prohibitive amounts of unsafe, interactive data collection in the real world. At the same time, bootstrapping from real-world demonstration data via inverse distillation makes the exploration problem tractable for RL fine-tuning in simulation. This minimizes the amount of task-specific engineering required by algorithm designers such as designing the dense rewards or manually designing the scenes.

Concretely, we propose \MethodName, a system for robustifying real-world imitation learning policies without requiring significant human effort, by constructing realistic simulation analogs for real-world environments on the fly and using these for robust policy learning. Our contributions include:

\begin{itemize}
    \item A simple policy learning pipeline that synthesizes controllers to perform diverse manipulation tasks in the real world that (i) reduces human effort in constructing environments and specifying rewards, (ii) produces robust policies that transfer to real-world,  cluttered scenes, showing robustness to disturbances and distractors, (iii) requires minimal amounts of expensive and unsafe data collection in the real world. 
    \item A novel algorithm for transferring demonstrations from the real world to the reconstructed simulation to bootstrap efficient reinforcement learning from the low-level Lagrangian state for policy fine-tuning. We show that this real-to-sim transfer of human demonstrations both improves efficiency and biases policies toward realistic behavior in simulation which effectively transfers back to the real world. 
    \item An intuitive graphical interface for quickly scanning and constructing digital twins of real-world scenes with articulation, separated objects, and accurate geometries. 
    \item We present extensive experimental evaluation showing that \MethodName produces reactive policies that solve several manipulation tasks in real-world scenes under physical disturbances and visual distractions. Across eight diverse tasks, our pipeline provides an improvement of 67$\%$ over baselines in average success rate across scenarios with varying object poses, visual distractors, and physical perturbations. 
\end{itemize}

\section{Related Work}
\label{sec:relatedwork}

\textbf{Learning Visuomotor Control from Demonstrations}:
Behavior cloning (BC) of expert trajectories can effectively acquire robot control policies that operate in the real world~\cite{florence2022implicit, chi2023diffusion, zhao2023aloha, brohan2022rt, florence2019self, mandlekar2021robomimic}. 
While several works have used BC to learn performant policies from small to moderately-sized datasets~\cite{chi2023diffusion, zhao2023aloha, mandlekar2021robomimic}, performance tends to drop when the policy must generalize to variations in scene layouts and appearance.
Techniques for improving BC often require much larger-scale data collection~\cite{brohan2022rt,ross2011dagger}, raising scalability concerns.
Other techniques support generalization with intermediate representations ~\cite{florence2019self} and leverage generative models to add visual distractors~\cite{yu2023scaling,mandi2022cacti}. These can improve robustness to visual distractors but do not address physical or dynamic disturbances, as these require producing actions not present in the data. 

\textbf{Fine-tuning Imitation with RL and Improving RL with Demonstrations}: 
Reinforcement learning has been used to improve the performance of models originally trained with imitation learning.
RL has exploded in its capacity for fine-tuning LLMs~\cite{ouyang2022training} and image generation models~\cite{black2023training}, learning rewards from human feedback \cite{christiano2017deep}.
In robotics, prior work has explored techniques such as offline RL \cite{yang2023robot, nair2020awac, kumar2020conservative}, learning world models \cite{mendonca2023structured,feng2023finetuning}, and online fine-tuning in the real world \cite{ball2023efficient,balsells2023autonomous,gupta2021reset,yang2023robot}. 
Expert demonstrations have also been used to bootstrap exploration and policy learning with RL~\cite{james2022coarse, james2022q, rajeswaran2017learning, zhu2019dexterous}. 
We similarly combine imitation and RL to guide exploration in sparse reward settings.
However, our pipeline showcases how demonstrations additionally benefit RL by biasing policies toward physically plausible solutions that compensate for imperfect physics simulation.

\textbf{Sim-to-real policy transfer}: 
RL in simulation has been used to synthesize impressive control policies in a variety of domains such as locomotion~\cite{margolis2022rapid, lee2020learning,kumar2021rma}, dexterous in-hand manipulation~\cite{chen2022system, chen2023visual, andrychowicz2020learning, handa2023dextreme}, and drone flight~\cite{song2023reaching}. Many simulation-based RL methods leverage some form of domain randomization~\cite{tobin2017domain, peng2018sim}, system identification~\cite{hwangbo2019learning, tan2018sim}, or improved simulator visuals~\cite{rao2020rlcyclegan, ho2021retinagan} to reduce the simulation-to-reality (sim-to-real) domain gap.
Prior work has also shown the benefit of ``teacher-student'' distillation~\cite{chen2023visual, kumar2021rma, shenfeld2023tgrl,chen2020learning}, wherein privileged ``teacher'' policies learned quickly with RL are distilled into ``student'' policies that operate on sensor observations.
To acquire transferable controllers, we similarly leverage GPU-accelerated simulation, teacher-student training, and domain randomization across parallel environments.
However, we address the more challenging scenario of household manipulation, which is characterized by richer visual scenes, and minimize the necessary engineering effort by relying on sparse rewards. We also simplify sim-to-real by training on digital twin assets and co-training with real data \cite{wang2024poco}.

\textbf{Real-to-sim transfer of scenes}: 
Designing realistic simulation environments has been studied from the perspective of synthesizing digital assets that reflect real objects.
Prior work has used tools from 3D reconstruction~\cite{jiang2022ditto} and inverse graphics~\cite{chen2023urdformer} for creating digital twins, and such real-to-sim pipelines have been used for both rigid and deformable~\cite{sundaresan2022diffcloud} objects.
These approaches are all compatible with our system and could be used to automate real-to-sim scene transfer and reduce human effort.
Our work similarly leverages advancements in 3D vision~\cite{tancik2023nerfstudio} for reconstructing object geometry, but we also introduce an easy-to-use GUI for building a URDF/USD with accurate articulations.
Furthermore, our GUI could be used to improve the aforementioned methods by making it easier to collect a large dataset of human-annotated articulated scenes. 
The accuracy of the simulator could be improved further combining our GUI with the latest system identification research \cite{memmel2024asid,lim2022planarrobotcastingreal2sim2real}.

\textbf{Real-to-sim-to-real transfer}: 
Prior work has used NeRF~\cite{mildenhall2021nerf} and other 3D reconstruction techniques to create realistic scene representations for improving manipulation ~\cite{zhou2023nerf}, navigation~\cite{deitke2023phone2proc,chang2023goat} and locomotion~\cite{byravan2023nerf2real}. These works, however, only use the visual component of the synthetic scene and do not involve any physical interaction with a reconstructed geometry. As a result, these systems cannot adjust to environmental changes beyond visual distractions. For instance, different grasp poses may require different placements, and a policy cannot discover these novel behaviors without physically interacting with the environment during training.
A limited number of works have learned policies that interact with the reconstructed environments, but they either simplify the reconstructed shapes \cite{liu2020real} or are limited to simple grasp motions \cite{wang2023real2sim2real}.

\section{\MethodName: A Real-to-Sim-to-Real System for Robust Robotic Manipulation}
\label{sec:method}
\begin{figure*}[h]
    \centering
    \includegraphics[width=0.9\linewidth]{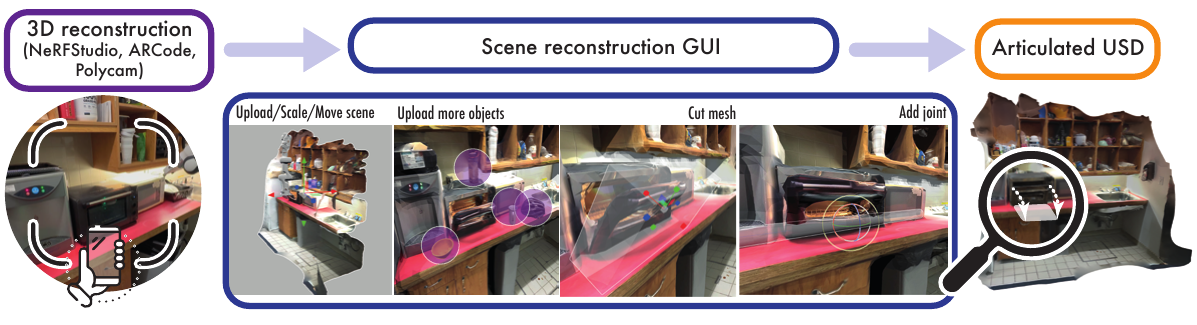}
    \caption{\footnotesize{Overview of the real-to-sim pipeline for transfering scenes to the simulator. The first stage consists of scanning the environment, using off-the-shelf tools such as NeRFStudio, ARCode, or Polycam. Each has its strengths and weaknesses and should be used appropriately (see Appendix \ref{apdx:r2simscenes} for recommendations). The second stage consists of uploading the reconstructed scene into \MethodName's GUI where the user can cut the mesh, specify joints, and organize the scene as desired. Once complete, the scene can be downloaded as a USD asset, which can be directly imported into the simulator. }}
    \label{fig:real2simpipeline-simgeneration}
\end{figure*}

\subsection{System Overview}
\label{sec:systemoverview}
Our goal is to obtain a control policy that maps real-world sensory observations to robot actions. We only assume access to a small set of demonstrations $(\sim 15)$ containing (observation, action) trajectories collected by an expert, although in principle \MethodName can also be used to robustify large, expressive pretrained models as well. Our approach robustifies real-world imitation learning policies using simulation-based RL to make learned controllers robust to disturbances and distractors not present in the demos. The proposed pipeline, \MethodName, achieves this with four main steps (Fig~\ref{fig:teaser}):
\begin{enumerate}[leftmargin=*]
    \item We construct geometrically, visually, and kinematically accurate simulation environments from real-world image capture. We leverage 3D reconstruction tools and develop an easy-to-use graphical interface for adding articulations and physical properties.
    \item We obtain a set of successful trajectories containing privileged information (such as Lagrangian state, e.g. object and joint poses) in simulation. We propose an ``inverse distillation'' algorithm to transfer a policy learned from real-world demonstrations to create a dataset of trajectories (i.e., demos) in the simulation environment.
    \item The synthesized simulation demos bootstrap efficient fine-tuning with RL in simulation using an \textit{easy-to-design} sparse reward function and low-dimensional state space, with added randomization to make the policy robust to environmental variations.
    \item The learned policy is transferred to reality by distilling a state-based simulation policy into a policy operating from raw sensor observations available in the real world~\cite{chen2020learning,chen2023visual}. During distillation, we also co-trained with the original real-world demonstrations to capitalize on the combined benefits of simulation-based robustification and in-domain real-world data. 
\end{enumerate}
The following sections describe each component in detail, along with a full system overview in Fig~\ref{fig:teaser}.

\begin{figure*}[h]
    \centering
    \includegraphics[width=0.85\linewidth]{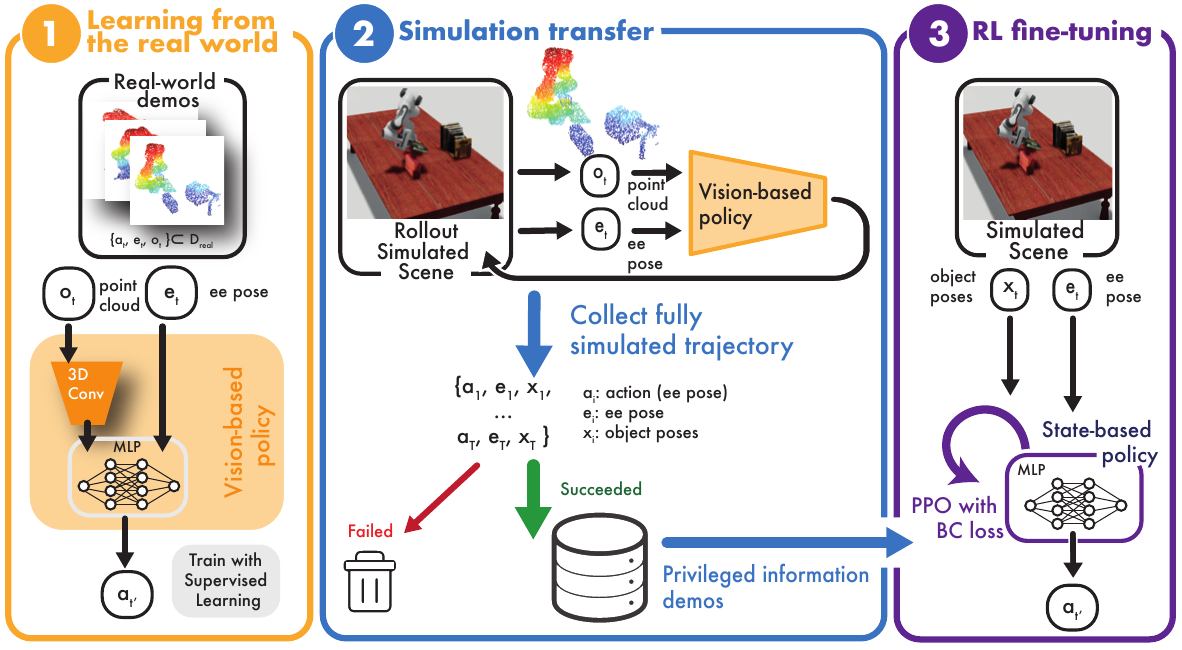}
    \caption{\footnotesize{\textbf{Inverse distillation \& RL fine-tuning}. We introduce a novel procedure for going from point cloud-based policies trained from real-world demonstrations $\mathcal{D}_{\text{real}}$ to a robust privileged state-based policy in simulation. 1) Train a vision-based policy with supervised learning on $\mathcal{D}_{\text{real}}$ 2) Rollout the vision-based policy on the simulation rendered point clouds and collect a set of 15 privileged demonstrations with object poses, $\mathcal{D}_{\text{sim}}$ 3) Train a robust state-based policy with RL and a sparse reward, adding a BC loss fitting $\mathcal{D}_{\text{sim}}$ to bias exploration and set a prior on real-world-transferable policies.}}
    \label{fig:real2sim-policy}
\end{figure*}

\subsection{Real-to-Sim Transfer for Scalable Scene Generation}
\label{sec:r2simscenes}

The first step of \MethodName is to construct geometrically, visually, and kinematically realistic simulated scenes for policy training. This requires (i) generating accurate textured 3D geometry from real-world images and (ii) specifying articulations and physical parameters. For geometry reconstruction, we use existing off-the-shelf 3-D reconstruction techniques. Our pipeline is agnostic to the particular method used, and we have verified the approach with a variety of scanning apps (e.g., Polycam~\cite{polycam2020} and ARCode~\cite{arcode2022}) and 3D reconstruction pipelines~\cite{tancik2023nerfstudio, mueller2022instant}, each of which convert a set of multi-view 2D images (or a video) into a textured 3D mesh. The raw mesh denoted $\mathcal{G}$, is typically exported as a single globally-unified geometry, which is unsuitable for direct policy learning. Scene objects are not separated and the kinematics of objects with internal joints are not reflected. Physical parameters like mass and friction are also required and unspecified. We therefore further process the raw mesh $\mathcal{G}$ into a set of separate bodies/links $\{\mathcal{G}_{i}\}_{i=1}^{M}$ with kinematic relations $\mathcal{K}$ and physical parameters $\mathcal{P}$.

\begin{figure}
    \centering
    \includegraphics[width=\linewidth]{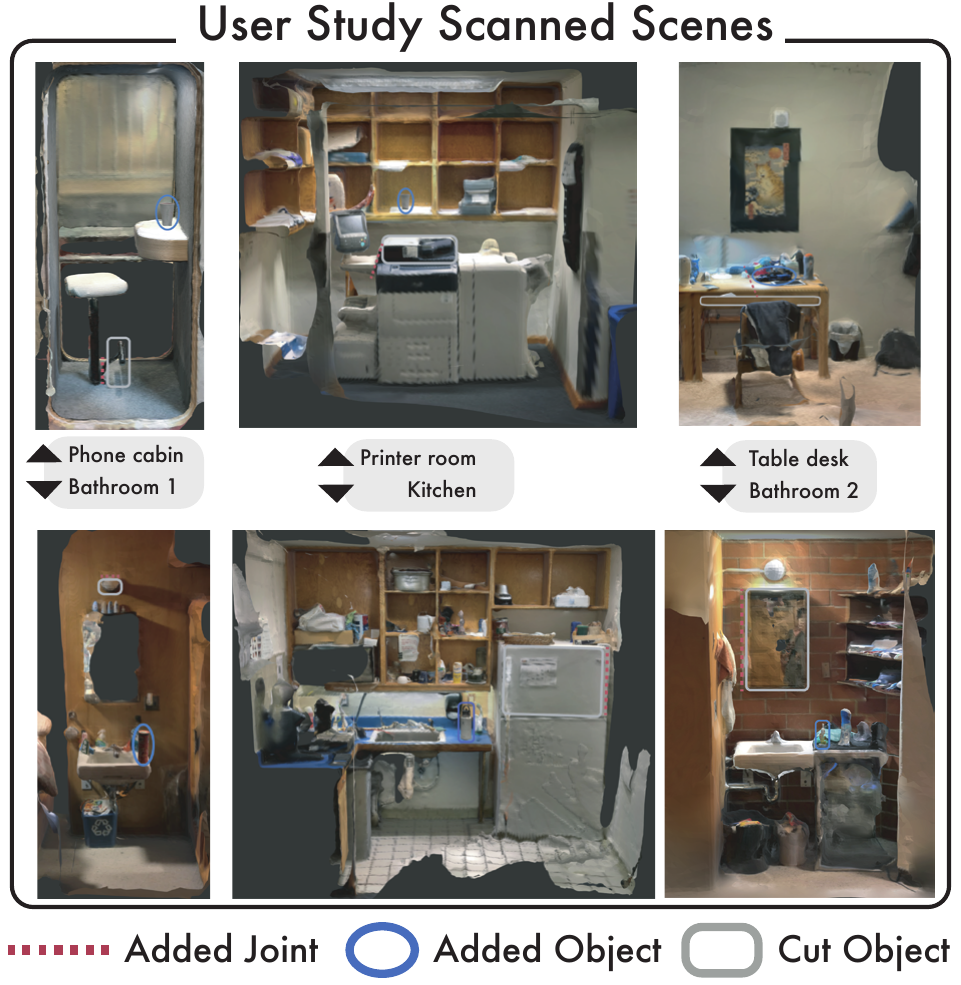}
    \caption{\footnotesize{Overview of the scenes assembled by the Users during the user study, see Section \ref{sec:userstudies}.}} 
    \label{fig:userstudymeshes}
\end{figure}

While there are various automated techniques for automatically segmenting and adding articulations to meshes~\cite{jiang2022ditto}, in this work, we take a simple human-centric approach. We offer a simple graphical interface for humans to quickly separate meshes and add articulations (see Fig. \ref{fig:real2simpipeline-simgeneration}). Our GUI allows users to upload their own meshes and drag/drop, reposition, and reorient them in the global scene. Users can then separate meshes and add joints between different mesh elements, allowing objects like drawers, fridges, and cabinets to be scanned and processed. Importantly, our interface is lightweight, intuitive, and requires minimal domain-specific knowledge. We conducted a study (Section~\ref{sec:userstudies}) evaluating six non-expert users' experiences with the GUI and found they could scan complex scenes and populate them with a couple of articulated objects in under 15 minutes of active interaction time. Examples of real-world environments with their corresponding digital twins are shown in Fig \ref{fig:taskoverview} and Appendix Fig. \ref{fig:userstudymeshes}. 

The next question is \textemdash how do we infer the physics parameters that faithfully replicate the real world? While accurately identifying physical parameters is possible, this can be challenging without considerable interaction~\cite{bohg2017interactive, xu2019densephysnet}. While adapting to dynamics variations is an important direction for future work, in this system we set a single default value for mass and friction uniformly across objects and compensate for the sim-to-real gap to actual real-world values by constraining the learned policy to be close to a small number of real-world demonstrations as discussed in Section~\ref{sec:robustification}.

This procedure produces a scene $\mathcal{S} = \{  \{\mathcal{G}_{i}\}_{i=1}^{M}, \mathcal{K}, \mathcal{P} \}$ represented in a USD/URDF file that references the separated meshes and their respective geometric ($\mathcal{G}_{i}\}_{i=1}^{M}$), kinematics ($\mathcal{K}$) and physical parameters ($\mathcal{P}$). This environment can subsequently be used for large-scale policy robustification in simulation.

\subsection{Robustifying Real-World Imitation Learning Policies in Simulation}
\label{sec:robustification}
Given the simulation environment generated in Section \ref{sec:r2simscenes}, the next step in \MethodName involves learning a robust policy in simulation that can solve desired tasks from a wide variety of configurations and environmental conditions. While this can be done by training policies from scratch in simulation, this is often a prohibitively slow process, requiring considerable manual engineering. Instead, we will adopt a fine-tuning-based approach, using reinforcement learning in simulation to fine-tune a policy initialized from a small number of expert demonstrations collected in the real world. Since training RL directly from visual observations is challenging, we would ideally like to finetune simulation policies that are based on a privileged Lagrangian state. However, real-world demonstrations do \emph{not} have access to the low-level state information in the environment. To enable the bootstrapping of RL finetuning in simulation from a privileged state using real-world demonstrations, we introduce a novel ``inverse distillation'' (Section~\ref{sec:inversedistill}) procedure that is able to take real-world demonstrations with only raw sensor observations and actions and transfer them to simulation demonstrations, complete with low-level privileged state information. These privileged information demonstrations can then be used to instantiate an efficient RL-based fine-tuning procedure (Section~\ref{sec:rlfinetune}) in simulation to massively improve policy robustness. 

\subsubsection{Inverse-distillation from Real-to-Sim for Privileged Policy Transfer}
\label{sec:inversedistill}
We assume a human provides a small number of demonstrations  in the real world $\mathcal{D}_{\text{real}} = \{(o_1^i, a_1^i), \dots, (o_H^i, a_H^i)\}_{i=1}^N$, where trajectories contain observations $o$ (3D point clouds) and actions $a$ (delta end-effector pose). Considering that simulation-based RL fine-tuning is far more efficient and performant when operating from a compact state representation~\cite{kumar2021rma, chen2022system} (see Section~\ref{subsec:rlvision}) and we wish to use real-world human demonstrations to avoid the difficulties with running RL from scratch (see Section~\ref{subsec:rlscratch}), we want to transfer our observation-action demonstrations from the real world to simulation in a way that allows for subsequent RL fine-tuning in simulation from compact state-based representations. This presents a challenge because we do \emph{not} have an explicit state estimation system that provides a Lagrangian state for the collected demonstrations in the real world. We instead introduce a procedure, called ``inverse-distillation'', for converting our real-world set of demonstrations into a set of trajectories in simulation that are paired with privileged low-level state information.

Given the demonstrations $\mathcal{D}_{\text{real}}$, we can naturally train a policy $\pi_{\text{real}}(a|o)$ on this dataset via imitation learning. ``Inverse distillation'' involves executing this perception-based learned policy $\pi_{\text{real}}(a|o)$ in simulation, based on simulated sensor observations $o$, to collect a dataset $\mathcal{D}_{\text{sim}} = \{(o_1^i, a_1^i, s_1^i) \dots, (o_H^i, a_H^i, s_H^i)\}_{i=1}^M$ of successful trajectories which contain privileged state information $s_t^i$. The key insight here is that while we do not have access to the Lagrangian state in the real-world demonstrations when a learned real-world imitation policy is executed from \emph{perceptual} inputs in simulation, low-level privileged Lagrangian state information can naturally be collected from the simulation as well since the pairing between perceptual observations and Lagrangian state is known apriori in simulation. Since the goal is to improve \emph{beyond} the real-world imitation policy $\pi_{\text{real}}(a|o)$, we can then perform RL fine-tuning, incorporating the privileged demonstration dataset $\mathcal{D}_{\text{sim}}$ into the training process, as discussed in the following subsection. 

\subsubsection{Reinforcement Learning Fine-tuning in Simulation}
\label{sec:rlfinetune}
Given the privileged information dataset $\mathcal{D}_{\text{sim}}$, and the constructed simulation environment the goal is to learn a robust policy $\pi^*_{\text{sim}}(a|s)$ using reinforcement learning. There are two key challenges in doing so in a \emph{scalable} way: (1) resolving exploration challenges with minimal reward engineering, and (2) ensuring the policy learns behaviors that will transfer to the real world. 
`
We find that both challenges can be addressed by a simple demonstration augmented reinforcement learning procedure~\cite{shenfeld2023tgrl,nair2018overcoming,rajeswaran2017learning}, using the Lagrangian state-based dataset $\mathcal{D}_{\text{sim}}$. To avoid reward engineering, we define a simple reward function that detects if the scene is in a desired goal state (detailed sparse reward functions used in each task in Appendix~\ref{appdx:taskdetails}). We build on the proximal policy optimization \cite{schulman2017proximal} algorithm with the addition of an imitation learning loss as follows (where $\hat{A}_t$ is the estimator of the advantage function at step $t$ \cite{schulman2017proximal}, and $V_{\phi}$ is the learned value function):

\begin{equation}
    \label{rl_finetune}
    \begin{split}
            \max_{\theta,\phi} \alpha \sum_{(s_t, a_t, r_t)\in \tau_{\pi_{\theta_{\text{old}}}}}\text{min} (\frac{\pi_\theta(a_t|s_t)}{\pi_{\theta_{\text{old}}(a_t|s_t)}}\hat{A}_t, \\
            \text{clip}(\frac{\pi_\theta(a_t|s_t)}{\pi_{\theta_{\text{old}}(a_t|s_t)}}, 1-\epsilon, 1+\epsilon)\hat{A}_t)  \\
            + \beta \sum_{(s_t, V_t^\text{targ})\in \tau_{\pi_{\theta_{\text{old}}}}} (V_\phi(s_t) - V_t^\text{targ})^2  \\
            + \gamma \sum_{(s_i, a_i)\in \mathcal{D}_{\text{sim}}} \frac{\pi_\theta(a_i|s_i)}{\sum_{a_c}\pi_\theta(a_c|s_i)} 
    \end{split}
\end{equation}

In addition to mitigating issues associated with exploration~\cite{nair2018overcoming,rajeswaran2017learning}, leveraging the additional imitation learning term in the objective helps bias the policy toward physically plausible, safe solutions that improve transfer of behaviors to reality. During this process, we can train the policy for robustness by randomizing initial robot/object/goal poses. Appendix \ref{appdx:taskdetails} contains complete details of our training procedure. The result is a robust policy $\pi^*_{\text{sim}}(a|s)$ operating from Lagrangian state that is successful from a wide variety of configurations and environmental conditions. 

\subsection{Teacher-Student Distillation with Co-Training on Real-World Data for Sim-to-Real Transfer}
\label{sec:teacherstudentcotraining}
In previous sections, we described a method for efficiently learning a robust policy $\pi^*_{\text{sim}}(a|s)$ in simulation using privileged state information. However, in the real world, this privileged information is unavailable. Policy deployment requires operating directly from sensory observations (such as point clouds) in the environment. To achieve this, we build on the framework of teacher-student distillation (with interactive DAgger labeling)\cite{ross2011dagger,chen2023visual} where the privileged information policy $\pi^*_{\text{sim}}(a|s)$ serves as a teacher and the perceptual policy $\pi^*_{\text{real}}(a|o)$ is the student\footnote{For the sake of this work, we will assume that the optimal actions for the student and teacher coincide, and there are no information gathering specific challenges induced by partial observability~\cite{shenfeld2023tgrl}}. Since there is inevitable domain shift between simulation and real domains, this training procedure can be further augmented by co-training the distillation objective with a mix of the original real-world demonstration data $\mathcal{D}_{\text{real}}$ and simulation data drawn from $\pi^*_{\text{sim}}(a|s)$ (via the DAgger objective~\cite{chen2023visual}). This results in the following co-training objective for teacher-student policy learning: 
\begin{equation}
    \label{eq_distill}
    \begin{split}
            \max_{\theta} \alpha \sum_{(s_i, o_i, a_i)\sim \tau_{\pi_{\theta}}} \frac{\pi_\theta(\pi_{\text{teacher}}(s_i)|o_i)}{\sum_{a_c}\pi_\theta(a_c|o_i)}  \\
            + \beta \sum_{(o_i, a_i)\in \mathcal{D}_{\text{real}}}\frac{\pi_{\theta}(a_i|o_i)}{\sum_{a_c}\pi_\theta(a_c|o_i)} 
    \end{split}
\end{equation}

Here the first term corresponds to DAgger training in simulation, while the second term co-trains on real-world expert data. This allows the policy to take advantage of small amounts of high-quality real-world data to bridge the perceptual gap between simulation and real-world scenes and improve generalization compared to only using the data from simulation. We empirically demonstrate (Section \ref{sec:teacherstudentcotraining}) that this significantly increases the resulting success rate in the real world. On a practical note, we refer the reader to Appendix~\ref{apdx:implementationdetails} for additional details on the student-teacher training scheme that enables it to be successful in the proposed problem setting.

\begin{figure*}[h]
    \centering
    \includegraphics[width=0.95\linewidth]{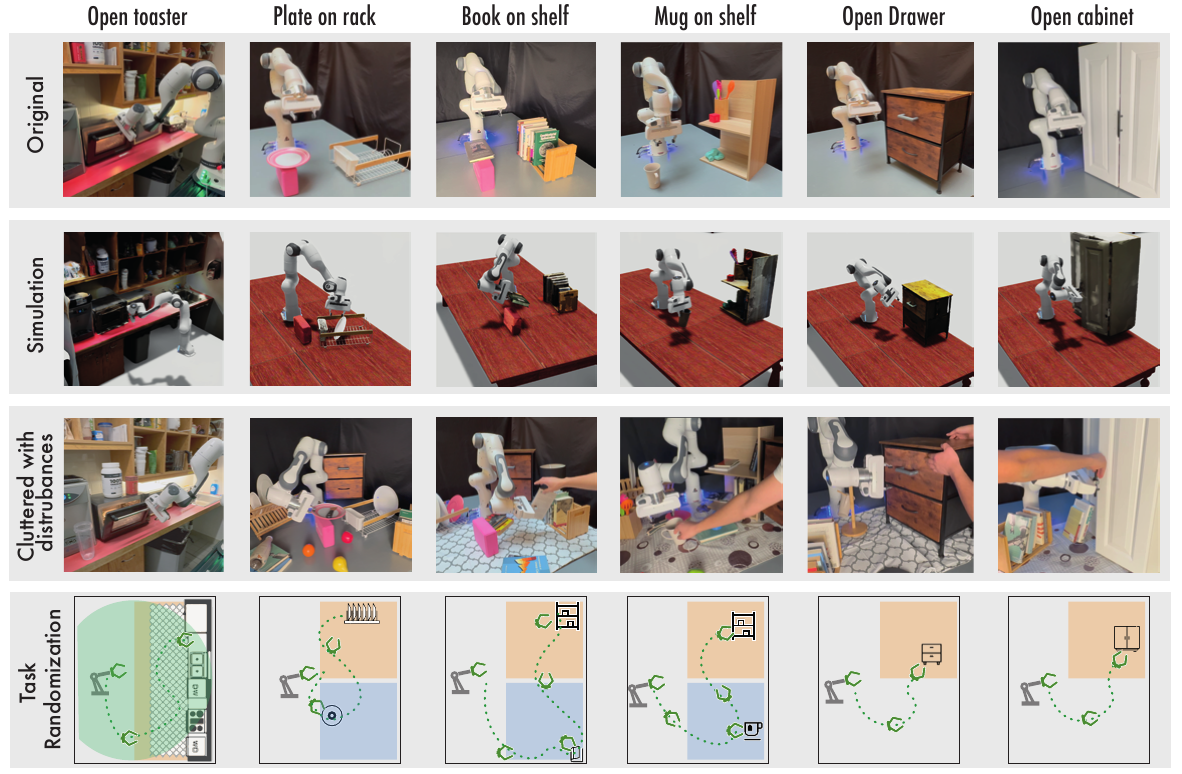}
    \caption{\footnotesize{ We depict the six tasks used to evaluate \MethodName. From top to bottom, we first show the original environment where we collect the demonstrations, second the simulated environment, third the environment where we do our final evaluation containing clutter and disturbances, and fourth the task randomization overview each shaded area corresponds to an approximation of how much randomization each object/robot have.}} 
    \label{fig:taskoverview}
\end{figure*}
\section{Experimental Evaluation}
\label{sec:experiments}

Our experiments are designed to answer the following questions about \MethodName:
(a) Does \MethodName provide real-world policies robust to variations in configurations, appearance, and disturbances?
(b) Does co-training policies with real-world data benefit real-world evaluation performance? 
(c) Is the real-to-sim transfer of scenes and policies necessary for training efficiency and the resulting performance?
(d) Does \MethodName scale up to more in-the-wild scenes?

To answer these questions, we evaluate \MethodName in eight different tasks, shown in Figure \ref{fig:taskoverview} and \ref{fig:inthewildenv}. These include 6-DoF grasping and reorientation of free objects (\emph{book on a shelf}, \emph{plate on a rack}, \emph{mug on a shelf}) and 6-DoF grasping and interacting with articulated objects (\emph{drawer} and \emph{cabinet}) on a tabletop and \emph{opening a toaster}, \emph{plate on a rack}, \emph{putting a cup in the trash} in more uncontrolled scenes. More details on the tasks such as their sparse reward functions and randomization setups are presented in Appendix \ref{appdx:taskdetails}. For each task, we consider three different disturbance levels in increasing order of difficulty (see Appendix \ref{appdx:taskdetails} for more details): 
\begin{enumerate}
    \item \textit{Randomizing object poses}: at the beginning of each episode we randomize the object and/or robot poses.
    \item \textit{Adding visual distractors}: at the beginning of each episode we also add visual distractors in a cluttered way. 
    \item \textit{Applying physical disturbances}: we apply physical disturbances throughout the episode rollout. We change the pose of the object being manipulated or the target location where the object needs to be placed, close the drawer/toaster/cabinet being manipulated, and move the robot base when possible.
\end{enumerate}
We conduct our experiments on a Franka Panda arm with the default parallel jaw gripper, using 6 DoF Cartesian end effector position control. For perceptual inputs, we obtain 3D point cloud observations from a single calibrated depth camera. More details on the hardware setup can be found in Appendix \ref{apdx:hardware}. All of the results in the real world are evaluated using the best policy obtained for each method, we report the average across at least 10 rollouts and the bootstrapped standard deviation. Videos of highlights and evaluation runs are available in the \href{https://real-to-sim-to-real.github.io/RialTo/}{website}. 

Throughout the next sections, we will evaluate \MethodName against the following set of baselines and ablations: 1) Imitation learning from 15 and 50 demos (Section~\ref{subsec:robustresults}); 2) No co-training on real-world data (Section~\ref{subsec:cotrainingresults}); 3) Co-training on demonstrations in simulation (Section~\ref{subsec:cotrainingresults}); 4) \MethodName from simulation demos (Section~\ref{subsec:r2simpolicies}); 5) Learning from an untargeted set of simulated assets (Section~\ref{subsec:r2simscenesbaseline}); 6) \MethodName without distractors (Section~\ref{subsec:rldistractors}); 7) \MethodName without demos (Section~\ref{subsec:rlscratch})

\subsection{\MethodName Learns Robust Policies via Real-to-Sim-to-Real}
\label{subsec:robustresults}
\begin{figure*}[h]
    \centering
    \includegraphics[width=0.95\linewidth]{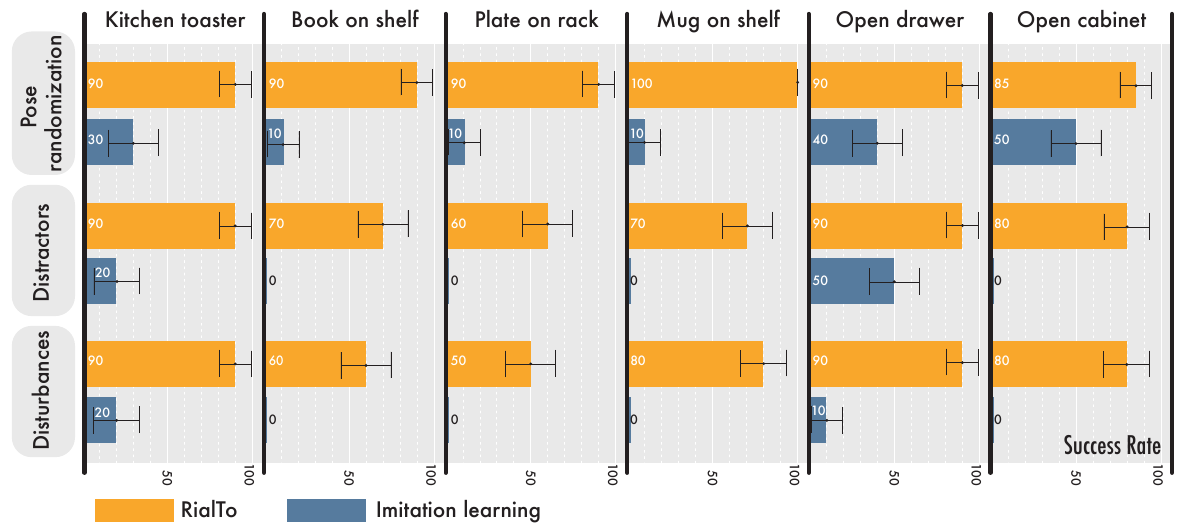}
    \caption{\footnotesize{Comparison of \MethodName against imitation learning both from 15 demonstrations. \MethodName provides robust policies across tasks and levels of distractions while imitation learning severely suffers when adding distractors and disturbances.}}
    \label{fig:mainres}
\end{figure*}

In this section, we aim to understand whether \MethodName can solve complex tasks, showing robustness to variations in configurations, disturbances, and distractors. We compare our approach of real-to-sim-to-real RL fine-tuning against a policy trained only on real-world demos via standard imitation learning (BC). We report the results of running \MethodName's pipeline starting from 15 demos collected directly in simulation and co-training with 15 real-world demos during the teacher-student distillation. In Section \ref{subsec:r2simpolicies} we show a comparison of running \MethodName uniquely on real or sim demos.

The results in Figure~\ref{fig:mainres} show \MethodName maintains high performance across configuration levels, achieving on average 91\% success across tasks for randomizing object poses, 77\% with distractors, and 75\% with disturbances. On the other hand, the presence of distractors and disturbances severely reduces the performance of pure imitation learning. For instance, when only randomizing the object poses, the BC baseline achieves an average of 25\% success rate across tasks. Under more challenging conditions, the BC baseline drops to 11\% and 5\% overall performance on average for distractors and disturbances, respectively. 

\begin{figure*}
    \centering
    \includegraphics[width=0.9\linewidth]{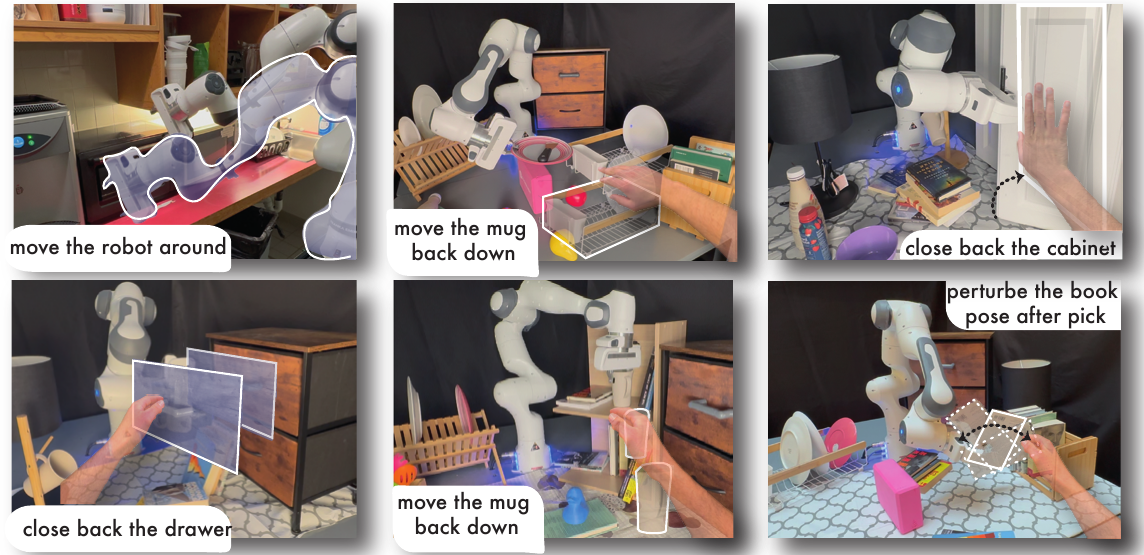}
    \caption{\footnotesize{Overview of the disturbances that \MethodName is robust to in the different tasks that we evaluated it on. }} 
    \label{fig:taskdistractors}
\end{figure*}

Figure~\ref{fig:teaser}, \ref{fig:taskdistractors} and the videos in the \href{https://real-to-sim-to-real.github.io/RialTo/}{website} qualitatively show how the resulting policies are robust to many kinds of environment perturbations, including moving the robot, moving the manipulated object and target positions, and adding visual distractors that cause occlusion and distribution shift. The policy rollouts also demonstrate error recovery capabilities, correcting the robot's behavior in closed loop when, e.g., objects are misaligned or a grasp must be reattempted. This highlights that \MethodName provides robustness that does not emerge by purely learning from demonstrations. 

\begin{table}[ht]
\begin{tabularx}{\linewidth}{@{}lc*{3}{>{\centering\arraybackslash}X}@{}}
\toprule
 & Only randomization & Distractors & Disturbances \\
\midrule
\addlinespace
BC (15 demos) & 10 $\pm$ 9\% & 0 $\pm$ 0\% & 0 $\pm$ 0\% \\
BC (50 demos) &  40 $\pm$ 15\% & 30 $\pm$ 16\% & 20 $\pm$ 13\% \\
\MethodName (15 demos) & \textbf{90} $\pm$ 9\%  & \textbf{70} $\pm$ 14\% & \textbf{60} $\pm$ 16\%\\

\bottomrule
\end{tabularx}
\caption{\MethodName and imitation learning on placing a book on the shelf.}
\label{tab:extrail}
\vspace{-15pt}
\end{table}

We also compare \MethodName against behavior cloning with 50 demonstrations to show that the problem is not simply one of slightly more data. Collecting the 50 demonstrations takes a total time of 1 hour and 45 minutes, which is significantly more than the human effort for \MethodName for which we collect 15 demos, in 30 minutes, and build the environment in 15 minutes of active time (see Section \ref{sec:userstudies}). Although more data improves the performance of direct imitation learning from 10\% to 40\%, 0\% to 30\%, and 0\% to 20\% for the three different levels of robustness, the results in Table~\ref{tab:extrail} show that \MethodName achieves approximately 2.5 times higher success rate than pure BC, despite using less than one third the number of demonstrations and taking less than half of the human supervision's time.

\subsection{Impact of Co-Training with Real-World Data}
\label{subsec:cotrainingresults}
\begin{figure}[t]
    \centering
    \includegraphics[width=\linewidth]{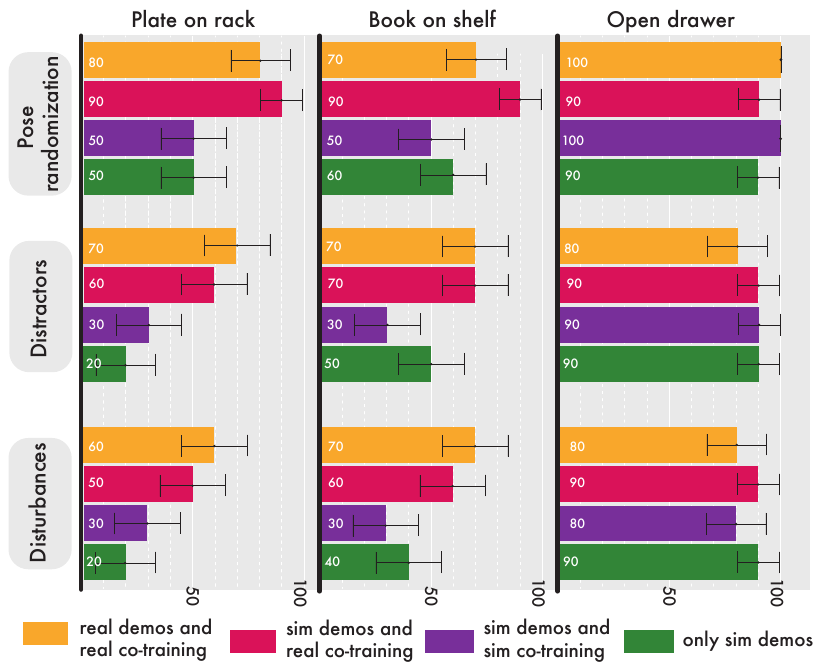}
    \caption{\footnotesize{Comparison between running \MethodName on sim vs real data. The performance on the methods doing co-training with real-world demos is higher than using only simulated demos or no real-world co-training, on the harder tasks( \emph{plate on rack} and \emph{book on shelf}), and matches the performance in the easier tasks( \emph{open drawer} and \emph{mug on shelf}). Furthermore, starting from real-world or simulated demos does equally well.}}
    \label{fig:cotrainingandrealvssim}
\end{figure}

Next, we assess the benefits offered by co-training with real-world demonstrations during teacher-student distillation, rather than just purely training policies in simulation. We consider the \emph{book on shelf}, \emph{plate on rack}, \emph{mug on shelf}, and \emph{open drawer} tasks (the two first being two of the harder tasks with lower overall performance). The results in Figure~\ref{fig:cotrainingandrealvssim} illustrate that co-training the policy with 15 real-world demonstrations significantly increases real-world performance on some tasks(3.5x and 2x success rate increase for \emph{book on shelf} and \emph{plate on rack} with disturbances, when comparing co-training on real-world demos against co-training with sim demos) while keeping the same performance on tasks that already have a small sim-to-real gap. Qualitatively, we observe the co-trained policy is more conservative and safer to execute. For instance, the policy without co-training usually comes very close to the plate or the book, occasionally causing it to fall. The policy with co-training data, however, leaves more space between the hand and the book before grasping, which is closer to the demonstrated behavior. The observation that sim co-training performs significantly worse than real-world co-training, indicates that co-training with real-world demonstrations is helping in reducing the sim-to-real gap for both the visual distribution shift between simulated and real point clouds and the sim-to-real dynamics gap.

\subsection{Is Real-to-Sim Transfer Necessary?}

\begin{figure}[h]
    \centering
    \includegraphics[width=0.9\linewidth]{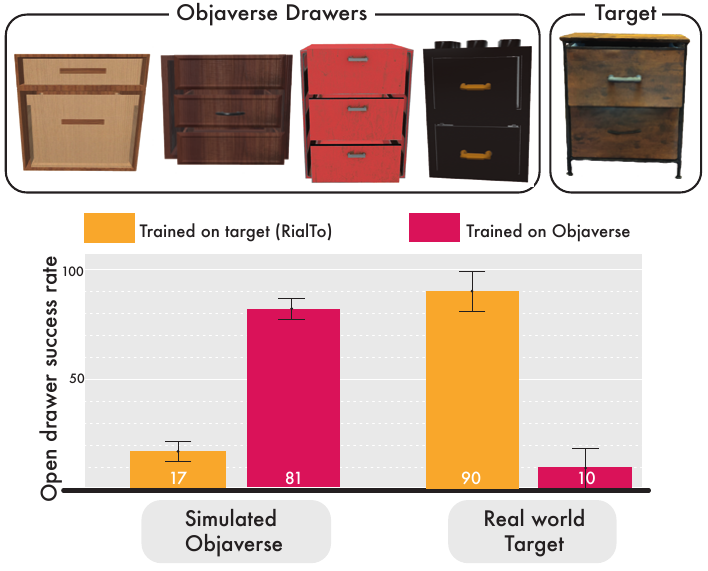}
    \caption{\footnotesize{Comparison between training with \MethodName on the reconstruction of the target drawer against training on a set of four drawers from Objaverse\cite{deitke2023objaverse}. We observe, that \MethodName on the real-to-sim asset does significantly better (90\% vs 10\%) when testing in the real world on the target drawer compared to training on the set of randomized drawers.}}
    \label{fig:simbaseline}
\end{figure}

\subsubsection{Real-to-Sim Transfer of Scenes}

\label{subsec:r2simscenesbaseline}
Instead of reconstructing assets from the target environment, one could train a policy on a diverse set of synthetic assets and hope the model generalizes to the real-world target scene~\cite{chen2023visual,gong2023arnold, wang2023gensim}. While this has shown promising results for object-level manipulation, such as in-hand reorientation ~\cite{chen2023visual}, it is still an active area of work for scene-level manipulation and rearrangement ~\cite{gong2023arnold}. Moreover, such methods require significant effort in creating a dataset of scenes and objects that enables the learned policies to generalize. Acquiring a controller that can act in many scenes is also a more challenging learning problem, requiring longer wall clock time, more compute, and additional engineering effort to train a performant policy on a larger and more diverse training set.

To probe the benefits of \MethodName over such a sim-only training pipeline, we compared the performance against a policy trained using only synthetic assets. Using an amount of time effort roughly comparable to what is required from a single user following our real-to-sim approach (see Section~\ref{sec:userstudies}), we collected a set of 4 drawers from the Objaverse dataset (see Figure \ref{fig:simbaseline}). Although this is small compared to the growing size of 3D object datasets, we found it non-trivial to transfer articulated objects into simulation-ready USDs and we leave it as future work. Given these manually constructed diverse simulation scenes, we then trained a multi-task policy using \MethodName from 20 demonstrations to open the 4 drawers. See Appendix \ref{apdx:simassetsbaseline} for the minor modifications to incorporate multi-task policy learning to \MethodName. 

As shown in Figure \ref{fig:simbaseline}, when evaluating the real target drawer, the policy trained on multiple drawers only achieves a 10\% success rate, much lower than the 90\% obtained by the policy trained on the target drawer in simulation. This leads us to conclude that to train a generalist agent, considerably more data and effort are needed as compared to the relatively simple real-to-sim procedure we describe for test time specialization. Moreover, this suggests that for performance on particular deployment environments, targeted generation of simulation environments via real-to-simulation pipelines may be more effective than indiscriminate, diverse procedural scene generation.

\subsubsection{Real-to-sim transfer of policies}
\label{subsec:r2simpolicies}
We additionally want to understand the impact of transferring policies from real-world demonstrations in comparison to running the pipeline starting with demos collected directly in simulation. This helps analyze whether instead of collecting demos both in simulation and in the real world (for the co-training) we can simply collect demos in the real world and do all the training with those.

Figure~\ref{fig:cotrainingandrealvssim} shows the real-world performance of policies trained using \MethodName when starting the RL fine-tuning step using real-world demonstrations as explained in \ref{sec:inversedistill} against using demonstrations provided directly in simulation. We observe that the performance for both cases is very close. These results show that \MethodName successfully learns policies with demonstrations from either source of supervision as long as we keep co-training the policies with real-world data in the teacher-student distillation step. Firstly, this indicates that we do \textbf{not} need to collect both demos in sim and real, but we can run \MethodName uniquely from the demos in the real world. Furthermore, this flexibility is a strength of our pipeline, as the ease of acquiring different sources of supervision may vary across deployment scenarios -- i.e., one could use policies pretrained from large-scale real-world data or obtain data from a simulation-based crowdsourcing platform.

\subsection{Scaling \MethodName to In-the-Wild Environments}
\label{sec:rialtowild}
\begin{figure}[h]
    \centering
    \includegraphics[width=\linewidth]{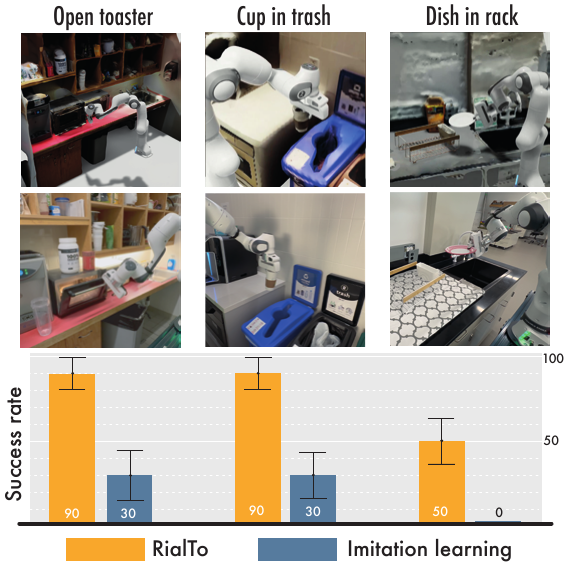}
    \caption{\footnotesize{We test \MethodName on uncontrolled and in-the-wild scenes, and we see we can continue to solve a variety of tasks more robustly than imitation learning techniques. } }
    \label{fig:inthewildenv}
\end{figure}

In this section, we scale up \MethodName to more uncontrolled and in-the-wild environments. We test \MethodName on three different tasks: \emph{open the microwave in a kitchen} (also shown in Section \ref{subsec:robustresults}), \emph{put a cup in the trash}, and \emph{bring the plate from the sink to the dishrack}. We observe that \MethodName scales up to these more diverse scenes and continues to perform significantly better than standard imitation learning techniques. In particular, \MethodName brings on average a 57\% improvement upon standard imitation learning, see Fig \ref{fig:inthewildenv}.
\section{Further Analysis and Ablations}
\label{sec:ablations}

\begin{table}[t!]
\centering
\begin{tabularx}{\linewidth}{@{}lc*{5}{>{\centering\arraybackslash}X}@{}}
\toprule
 & Pose Randomization & Distractors\\
\midrule
\addlinespace
\MethodName without distractor training & 60 $\pm$ 15\% & 30 $\pm$ 15\% \\
\MethodName with distractor training &  \textbf{100} $\pm$ 0\% & \textbf{70} $\pm$ 15\% \\
\bottomrule
\end{tabularx}
\caption{Real-World performance of policies trained with and without distractors on the task of placing a mug on a shelf.}
\label{tab:distractors}
\vspace{-15pt}
\end{table}

\subsection{Training with Distractors}
\label{subsec:rldistractors}
When performing teacher-student distillation we performed randomization with additional visual distractors to train a more robust policy that succeeds even in visual clutter. We analyze how this affects the final robustness of the learned policy. For the sake of analysis, we consider the performance on the \emph{mug on the shelf} task. The small size of the mug and its resemblance in shape and size to other daily objects make the visual component of this task particularly challenging when other objects are also present. 
Our findings in Table \ref{tab:distractors} show that adding distractors during training increases the success rate from 30\% to 70\% when testing the policy in environments with distractors. We also observe a performance improvement in setups with no distractors suggesting that such training also supports better sim-to-real policy transfer. 

\subsection{Comparison to RL from Scratch}
\label{subsec:rlscratch}
\begin{table*}[t!]
\centering
\begin{tabularx}{\textwidth}{@{}lc*{5}{>{\centering\arraybackslash}X}@{}}
\toprule
 & Open & Book on & Plate on & Mug on & Open \\ 
 & toaster & shelf & rack & shelf & drawer  \\ 
\midrule
\addlinespace
RL from scratch with 0 demos & 62 $\pm$ 2\% & 0 $\pm$ 0\% & 2 $\pm$ 0\% & 0 $\pm$ 0\% & 0 $\pm$ 0\% \\ 
RL fine-tuning from 15 real demos & 91 $\pm$ 1\% & \textbf{90} $\pm$ 1\% & \textbf{81} $\pm$ 2\% & \textbf{81} $\pm$ 2\% & \textbf{96} $\pm$ 1\% \\ 
RL fine-tuning from 15 sim demos & \textbf{96} $\pm$ 1\% & \textbf{89} $\pm$ 1\% & \textbf{82} $\pm$ 2\% & \textbf{82} $\pm$ 2\% & \textbf{95} $\pm$ 1\% \\ 

\bottomrule
\end{tabularx}
\caption{Comparison of training RL from scratch against RL from real and sim demos. RL from sim and real demos seem to be equivalent in most cases, but RL from scratch barely solves the task.}
\label{tab:r2simres}
\vspace{-15pt}
\end{table*}

We hypothesize two key advantages of incorporating demonstrations in the finetuning process: (1) aiding exploration, and (2) biasing the policy toward behaviors that transfer well to reality. Results in Table~\ref{tab:r2simres} show that training from PPO from scratch fails (0$\%$ success) in three out of five tasks and much poorer performance in the other two tasks. On tasks with non-zero success, we observed that the policy exploits simulator inaccuracies and learns behaviors that are unlikely to transfer to reality. (see Appendix Fig. \ref{fig:rlscratchapdx}). For example, the PPO policy opens the toaster by pushing on the bottom of the toaster, leveraging the slight misplacement of the joint on the toaster. Such behaviors are unsafe and would not transfer to reality, underlining the importance of using demonstrations during policy robustification.

\subsection{RL from Vision}
\label{subsec:rlvision}
\MethodName's ``inverse distillation" procedure to a compact state-space adds some methodological overhead to the system when compared to the possibility of doing RL fine-tuning directly on visual observations. However, as reported in Appendix Fig. \ref{fig:visionrlbaseline}, on the task of drawer opening, RL from compact states achieves a 96\% success rate after 12 hours of wall-clock time, while RL from vision only achieves a 1\% success rate after 35 hours. Hence, inverse distilling to state space is necessary because training RL from vision with sparse rewards is prohibitively slow, motivating the methodology outlined in Section~\ref{sec:inversedistill}.
\section{User study}
\label{sec:userstudies}

\begin{figure}[h]
    \centering
    \includegraphics[width=\linewidth]{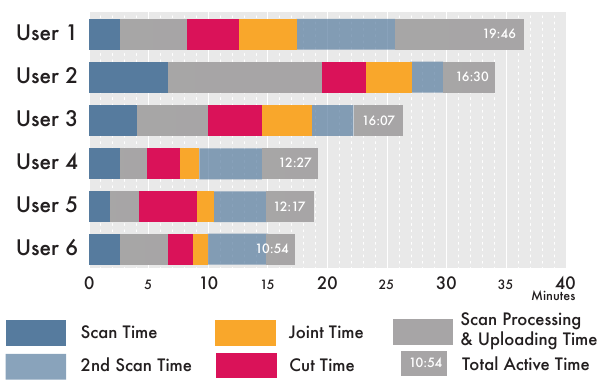}
    \caption{\footnotesize{3D reconstruction GUI's user study breakdown times. On average it takes 14 minutes and 40 seconds of active time or 25 minutes and 12 seconds of total time to create a scene through our proposed pipeline.}}
    \label{fig:userstudy}
\end{figure}

We analyzed the usability of \MethodName's pipeline for bringing real-world scenes to simulation. We ran a user study over 6 people, User 6 being an expert who used the GUI before and Users 1-5 never did any work on simulators 
before. Each participant was tasked with creating an articulated scene using the provided GUI. More precisely, their task was to: 1) scan a big scene, 2) cut one object, 3) scan and upload a smaller object, and 4) add a joint to the scene. From Figure \ref{fig:userstudy}, we found that the average total time to create a scene was 25 minutes and 12 seconds of which only 14 minutes and 40 seconds were active work. We also observed that the expert user accomplished the task faster than the rest, and twice as fast as the slowest user. This indicates that with practice, our GUI allows users to become faster at generating scenes. We conclude that doing the real-to-sim transfer of the scenes using the proposed GUI seems to be an intuitive process that is neither time nor labor-intensive when compared to collecting many demonstrations in the real world. We provide more details about the study in Appendix \ref{apdx:userstudy}.

\section{Limitations and Conclusion}
\label{sec:conclusion}
\paragraph*{Limitations} While our use of 3D point clouds instead of RGB enables easier sim-to-real transfer, we require accurate depth sensors that can struggle to detect thin, transparent, and reflective objects. Future work may investigate applying \MethodName to train policies that operate on RGB images or RGBD, as our framework makes no fundamental assumptions that prevent using different sensor modalities. We are also limited to training policies for tasks that can be easily simulated and for real-world objects that can be turned into digital assets. Currently, this is primarily limited to articulated rigid bodies, but advancements in simulating and representing deformables should allow our approach to be applied to more challenging objects. Even though we show \MethodName works on fast controllers, these are still relatively slow to minimize the sim-to-real gap in dynamics, thereafter there is potential to investigate tasks for which faster controllers are needed. In this work, we consider relatively quasistatic problems, where exact identification of physics parameters is not necessary for the constructed simulation. This will become important as more complex environments are encountered. Finally, as we explain in Section \ref{apdx:compute}, \MethodName currently takes around 2 days of wall-clock time end-to-end to train a policy for each task, this time bottleneck makes continual learning infeasible and understanding how to obtain policies faster with minimal human supervision would be valuable. We expect with more efficient techniques for learning with point clouds and better parallelization, this procedure can be sped up significantly. 

\paragraph*{Conclusion} This work presents \MethodName, a system for acquiring policies that are robust to environmental variations and disturbances on real-world deployment. Our system achieves robustness through the complementary strengths of real-world imitation learning and large-scale RL on digital twin simulations constructed on the fly. Our results show that by importing 3-D reconstructions of real scenes into simulation and collecting a small amount of demonstration data, non-expert users can program manipulation controllers that succeed under challenging conditions with minimal human effort, showing enhanced levels of robustness and generalization.

\section*{Acknowledgments}
The authors would like to thank the Improbable AI Lab and the WEIRD Lab members for their valuable feedback and support in developing this project. In particular, we would like to acknowledge Antonia Bronars and Jacob Berg for helpful suggestions on improving the clarity of the manuscript, and Marius Memmel for providing valuable insights on learning from point clouds in the early stages of the project. This work was partly supported by the Sony Research Award, the US Government, and Hyundai Motor Company.

\hfill \break
\textbf{Author Contributions}

\textbf{Marcel Torne} conceived the overall project goals, investigated how to obtain real-to-sim transfer of scenes and policies, and robustly do sim-to-real transfer of policies, wrote
all the code for the policy learning pipeline and \MethodName's GUI for real-to-sim transfer of scenes, ran simulation and real-world experiments, wrote the paper, and was the primary author of the paper.

\textbf{Anthony Simeonov} helped with setting up the robot hardware, made technical suggestions on learning policies from point clouds, helped with the task of placing the plate on the rack, and actively helped with
writing the paper.

\textbf{Zechu Li} assisted in the early stage of conceiving the project and helped develop \MethodName's GUI for the real-to-sim transfer of the scenes.

\textbf{April Chan} led the user study experiments to analyze \MethodName's GUI.

\textbf{Tao Chen} provided valuable insights and recommendations on sim-to-real transfer.

\textbf{Abhishek Gupta} was involved in conceiving the goals of the project, assisted with finding the scope of the paper, suggested baselines and ablations, played an active role in writing the paper and co-advised the project.

\textbf{Pulkit Agrawal} suggested the idea of doing real-to-sim transfer of scenes, was involved in conceiving the goals of the project, suggested baselines and ablations, helped edit the paper, and co-advised the project.

\bibliographystyle{plainnat}
\bibliography{references}

\clearpage
\newpage
\newpage
\newpage
Next, we provide additional details of our work. More concretely:
\begin{itemize}
    \item \textbf{Task Details} \ref{appdx:taskdetails}: provides more details on the tasks used to evaluate \MethodName and the baselines.
    \item \textbf{Implementation Details} \ref{apdx:implementationdetails}: provides more detailed information on the exact hyperparameters such as network architectures, point cloud processing, and dataset sizes using in \MethodName.
    \item \textbf{Further Analysis} \ref{apdx:furtheranalysis}: we provide further details on \MethodName, more concretely on running RL from vision, RL from scratch and on the sim-to-real gap.
    \item \textbf{Hardware Setup} \ref{apdx:hardware}: Details on the robot hardware and cameras used for the experiments.
    \item \textbf{GUI for Real-to-Sim Transfer of Scenes} \ref{apdx:r2simscenes}: We provide further details on the GUI that we proposed together with advice on which scanning methods to use for each scenario.
    \item \textbf{GUI User Study} \ref{apdx:userstudy}: We explain how we ran the User Study together with visualizations of the scanned scenes.
    \item \textbf{Compute Resources}\ref{apdx:compute}: We give details on the compute used to run the experiments.
    
\end{itemize}
\section{Task details}
\label{appdx:taskdetails}

\begin{figure}
    \centering
    \includegraphics[width=\linewidth]{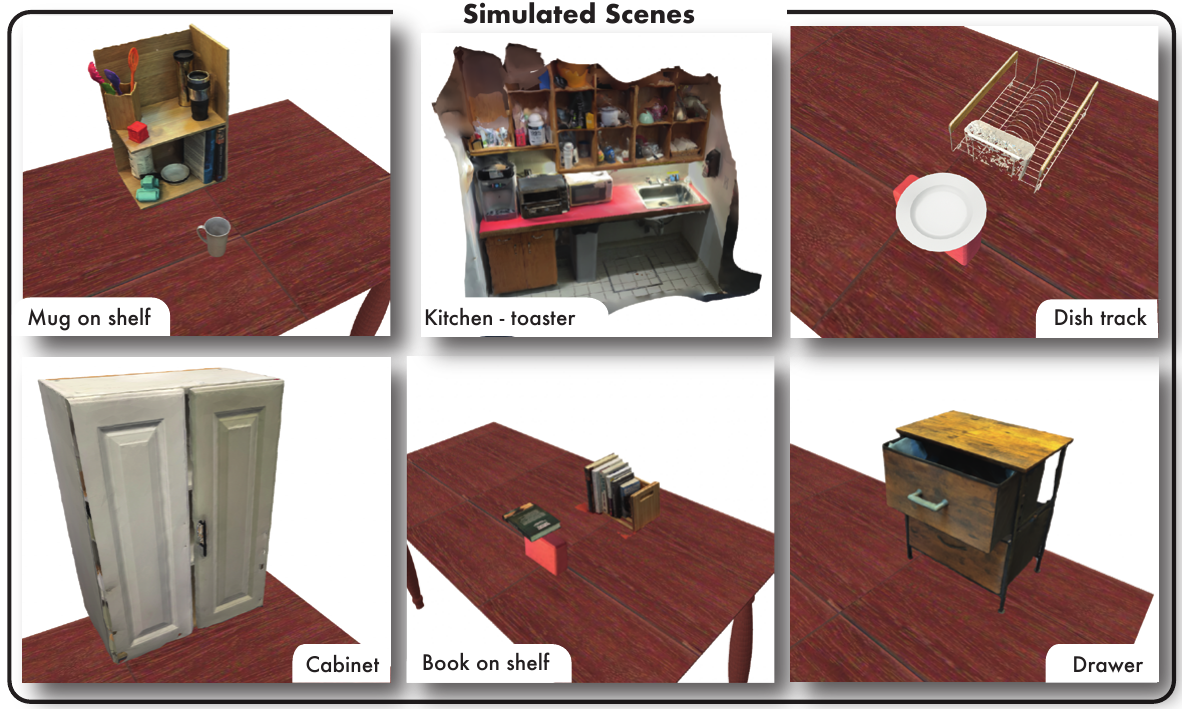}
    \caption{\footnotesize{Overview of the scenes generated using our GUI and used for evaluating \MethodName.}} 
    \label{fig:taskmeshes}
\end{figure}
In this section of the appendix, we describe additional details about each task. Across tasks, the state space consists of a concatenation of all of the poses of the objects present in the scenes together with the states of the joints and the state of the robot. The action space consists of a discretized end-effector delta pose of dimension 14. More concretely, we have 6 actions for the delta position, which moves $\pm 0.03$ meters in each axis, 6 more actions for rotating $\pm 0.2$ radians in each axis, and 2 final actions for opening and closing the gripper. 

As we explain in Section \ref{sec:r2simscenes}, we define a success function that will be used for selecting successful trajectories in the inverse distillation procedure and as a sparse reward in the RL fine-tuning phase. Next, we specify which are the success functions for each of the tasks:

\begin{itemize}
    \item \textbf{Kitchen Toaster}: $ success =$ \\$\text{toaster\_joint}>0.65 ~\&\&~ \text{condition(gripper\_open)}$
    \item \textbf{Open Drawer}: $ success =$ \\$\text{drawer\_joint}>0.1 ~\&\&~ \text{condition(gripper\_open)}$
    \item \textbf{Open Cabinet}: $ success =$ \\$\text{cabinet\_joint}>0.1 ~\&\&~ \text{condition(gripper\_open)}$
    \item \textbf{Plate on the rack}: $ success =$ \\$||\text{plate\_site}-\text{rack\_site}||_2<0.2 ~\&\&~ \text{rack\_y\_axis}\cdot\text{plate\_z\_axis} > 0.9 ~\&\&~ \text{condition(gripper\_open)}$
    \item \textbf{Book on shelf}: $ success =$ \\$||\text{book\_site}-\text{shelf\_site}||_2<0.12 \\~\&\&~ \text{condition(gripper\_open)}$
    \item \textbf{Mug on shelf}: $ success =$ \\$||\text{mug\_site}-\text{shelf\_site}||_2<0.12 ~\&\&~ \text{mug\_z\_axis}\cdot\text{shelf\_z\_axis} > 0.95~\&\&~\text{condition(gripper\_open)}$
    \item \textbf{Plate on the rack in the kitchen}: $ success =$ \\$||\text{plate\_site}-\text{rack\_site}||_2<0.2 ~\&\&~ \text{rack\_y\_axis}\cdot\text{plate\_z\_axis} > 0.9 ~\&\&~ \text{condition(gripper\_open)}$
    \item \textbf{Cup in trash}: $ success =$ $||\text{cup\_site}-\text{trash\_site}||_2<0.07 \\ ~\&\&~ \text{condition(gripper\_open)}$
    
\end{itemize}

\begin{table*}
\centering
\begin{tabularx}{\textwidth}{@{}lc*{6}{>{\centering\arraybackslash}X}@{}}
\toprule
 Task  & USD  Name & Episode length & Randomized  & Position  & Position  & Orientation & Orientation    \\
  Parameters &  &  & Object Ids &  Min (x,y,z) &  Max (x,y,z) & Min (z-axis) & Max (z-axis)  \\
 \midrule
\addlinespace
 Kitchen toaster & kitchentoaster3.usd & 130 & [267] & [0.3,-0.2,-0.2] & [0.7,0.1,0.2] & [-0.1] & [0.1]\\
 Plate on rack & dishinrackv3.usd & 150 & [278, [270,287]]  & [-0.4,-0.035,0] & [0,0.25,0] & [-0.52,0] & [0.52,0]\\
 Mug on shelf & mugandshelf2.usd & 150 & [267,263] & [[-0.3,0,0], [-0.1,0.25,0]] & [[0.25,0.3,0.07], [0.4,0.4,0]] & [-0.52,-0.54] & [0.52, 0.54]\\
 Book on shelf & booknshelve.usd & 130 & [277, [268,272]] & [[-0.25,-0.12,0], [-0.15,-0.05,0]] & [[0.15,0.28,0], [0.15,0.15,0]] & [-0.52,0] & [0.52,0] \\
 Open cabinet & cabinet.usd & 90 & [268] & [-0.5,-0.1,0.1] & [0,0.3,-0.1] & [-0.52] & [0.52]\\
 Open drawer & drawerbiggerhandle.usd & 80 & [268]& [-0.26,-0.07,-0.05] & [0.16,0.27,0] & -0.5 & 0.5\\
 Cup in trash & cupntrash.usd & 90 & [263, 266]& [[[-0.2, -0.3, -0.2], [-0.2,-0.12,0]]] & [[[0.2, 0.1, 0.2], [0.2,0.2,0]]] & [0,0] & [0,0]\\
 Plate on rack from kitchen & dishsinklab.usd & 110 & [[263, 278, 270]]& [[[-0.25, -0.1, -0.1], [-0.1,0.05,0], [-0.2,0,0]]] & [[[0.1, 0.2, 0.1], [0.1,0.15,0], [0,0,0]]] & [0,-0.3,0] & [0,0.3,0]\\
 
\midrule
\addlinespace

\bottomrule
\end{tabularx}
\caption{Specific parameters for each one of the tasks.}
\label{tab:paramstask}
\end{table*}
\begin{table*}
\centering
\begin{tabularx}{\textwidth}{@{}lc*{6}{>{\centering\arraybackslash}X}@{}}
\toprule
 Task & Position (x,y,z) & Rotation (quat) & Crop Min & Crop Max & Size    \\
  Parameters & Camera   & Camera  & Camera  & Camera  & Image \\
 \midrule
\addlinespace
 Kitchen toaster & [0.0, -0.37, 0.68] & [0.82,0.34,-0.20, -0.41] & [-0.8,-0.8,-0.8] & [0.8,0.8,0.8] & (640,480)\\
 Plate on rack & [0.95,-0.4,0.68] & [0.78,0.36, 0.21, 0.46] & [-0.3,-0.6,0.02] & [0.9,0.6,1] & (640,480) \\
 Mug on shelf & [0.95,-0.4,0.68] & [0.78,0.36, 0.21, 0.46] & [-0.3,-0.6,0.02] & [0.9,0.6,1] & (640,480) \\
 Book on shelf & [0.95,-0.4,0.68] & [0.78,0.36, 0.21, 0.46] & [-0.3,-0.6,0.02] & [0.9,0.6,1] & (640,480) \\
 Open cabinet & [0.95,-0.4,0.68] & [0.78,0.36, 0.21, 0.46] & [-0.3,-0.6,0.02] & [0.9,0.6,1] & (640,480) \\
 Open drawer & [0.95,-0.4,0.68] & [0.78,0.36, 0.21, 0.46] & [-0.3,-0.6,0.02] & [0.9,0.6,1] & (640,480) \\
 Cup in trash & [0.0, -0.37, 0.68] & [0.82,0.34,-0.20, -0.41] & [-1,-1,-1] & [1,1,1] & (640,480)\\
 Plate on rack from kitchen & [0.0, -0.37, 0.68] & [0.82,0.34,-0.20, -0.41] & [-0.8,-0.8,-0.8] & [0.8,0.8,0.8] & (640,480)\\
 
\midrule
\addlinespace

\bottomrule
\end{tabularx}
\caption{Camera parameters for each task.}
\label{tab:paramstask2}
\end{table*}

\subsection{Simulation details}

For simulating each one of the tasks, we use the latest simulator from NVIDIA, IsaacSim \cite{isaacsim2022}. Furthermore, to develop our code we were inspired by the Orbit codebase \cite{mittal2023orbit}, one of the first publicly available codebases that run Reinforcement Learning and Robot Learning algorithms on Isaac Sim. 

Regarding the simulation parameters of the environments, as mentioned in the text, we set default values in our GUI and these are the same that are used across the environments. In more detail, we use convex decomposition with 64 hull vertices and 32 convex hulls as the collision mesh for all objects. These values could vary in some environments, but we have found they are in general a good default value. There is one exception, the dish on the rack task, where the rack needs to be simulated very precisely, in that case, we used SDF mesh decomposition with 256 resolution which returns high-fidelity collision meshes. Note that all these options can be changed from our GUI. Regarding the physics parameters, we set the dynamic and static frictions of the objects to be 0.5, the joint frictions to be 0.1, and the mass of the objects to be 0.41kg. Note that in many of the tasks, we also leverage setting fixed joints on the objects, to make sure these won't move, for example, on the shelf or kitchen.

\section{Implementation details}
\label{apdx:implementationdetails}

\subsection{Network architectures}

\subsubsection{State-based policy}
As described in Section \ref{sec:rlfinetune}, we fine-tune a state-based policy with privileged information in the simulator. This policy is a simple Multi-Layer Perceptron (MLP) with two layers of size 256 each. This takes as input the privileged state from the simulator and outputs a Categorical distribution of size 14 encoding the probabilities for sampling each discrete end-effector action. For our PPO with BC loss implementation, we build on top of the Stable Baselines 3 repository \cite{stable-baselines3}. The network for the value function shares the first layer with the actor. See Table \ref{tab:ppoparams} for more details.
\begin{table*}
\centering
\begin{tabularx}{\textwidth}{@{}lc*{6}{>{\centering\arraybackslash}X}@{}}
\toprule
  MLP layers & PPO n\_steps & PPO batch size & PPO BC batch size & PPO BC weight & Gradient Clipping\\
\midrule
\addlinespace
 256,256 & episode length & 31257 & 32 & 0.1 & 5\\

\bottomrule
\end{tabularx}
\caption{State-based policy training parameters. The rest of the parameters are the default as described in Stable Baselines 3\cite{stable-baselines3}. }
\label{tab:ppoparams}
\end{table*}

\subsubsection{Point cloud policy}

For both the inverse distillation procedure (Section \ref{sec:inversedistill}) and the last teacher-student distillation steps (Section \ref{sec:teacherstudentcotraining}) we train a policy that takes as input the point cloud observation together with the state of the robot (end-effector pose and state) and outputs a Categorical distribution of size 14 encoding the probabilities for each action. The network architecture consists of an encoder of the point clouds that maps to an embedding of size 128. Then this embedding is concatenated to the state of the robot (size 9) and is passed through an MLP of size 256,256. Regarding the point cloud encoder, we use the same volumetric 3D point cloud encoder proposed in Convolutional Occupancy Networks \cite{peng2020convolutional}, consisting of a local point net followed by a 3D U-Net which outputs a dense voxel grid of features. These features are then pooled with both a max pooling layer and an average pooling layer and the resulting two vectors are concatenated to obtain the final point cloud encoding of size 128.

\subsection{Teacher-student distillation}
\label{appdx:teachestudentdistillation}
\begin{figure}
    \centering
    \includegraphics[width=\linewidth]{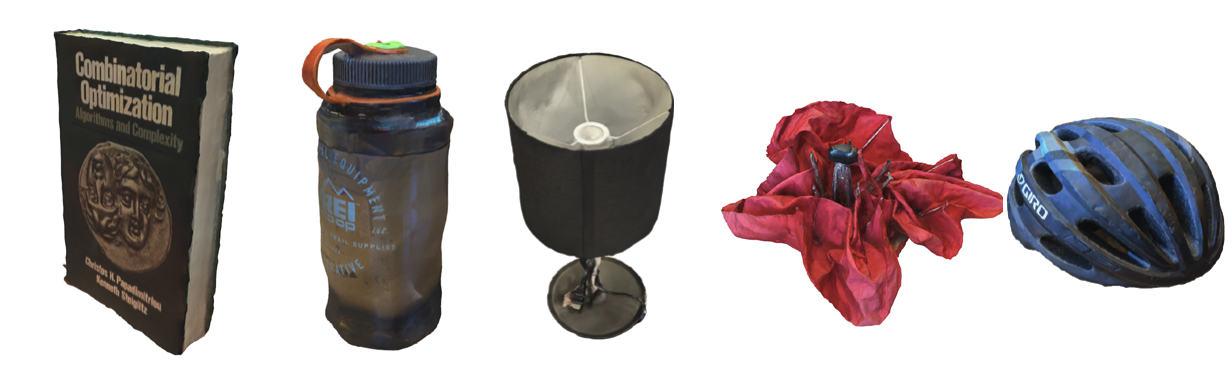}
    \caption{\footnotesize{Distractor objects used to get a robust policy to visual distractors in the teacher-student distillation step \ref{sec:teacherstudentcotraining}. }}
    \label{fig:distractors}
\end{figure}

Given the state-based policy $\pi_\text{sim}(a|s)$ learned in the simulator, we wish to distill it into a policy $\pi^*_\text{sim}(a|o)$ that takes the point cloud observation and outputs the action. We take the standard teacher-student distillation approach \cite{kumar2021rma, chen2023visual}. The first step consists of doing imitation learning on a set of trajectories given by the expert policy $\pi_\text{sim}(a|s)$ rollout. This set of trajectories needs to be carefully designed to build an implicit curriculum so that we can learn the student policy successfully. When designing this dataset of trajectories, we mix 15000 trajectories rendering full point clouds (where all faces of the objects are visible, which is obtained through directly sampling points from the mesh, as proposed in \cite{chen2023visual}), 5000 trajectories rendered from a camera viewpoint that is approximately the same position as the camera in the real world, a set of 2000 trajectories also generated from the same camera viewpoint in sim but adding distractor objects (see Figure \ref{fig:distractors}), finally, we mix the 15 real-world trajectories. The four different splits in the dataset are sampled equally, with 1/4 probability each.

After this first distillation step, we perform a step of DAgger \cite{ross2011dagger}, where we roll out the policy $\pi^*_\text{sim}(a|o)$ and relabel the actions with $\pi_\text{sim}(a|s)$. In this second and last step, we mix the DAgger dataset with the trajectories with distractors in sim and the real-world trajectories and sample trajectories. Again each dataset is sampled equally with 1/3 probability each. 

Finally, the details for generating and randomizing the point clouds are available in Table \ref{tab:pcddetails} and were largely inspired by \cite{chen2023visual}. The parameters for training the point cloud-based network are available in \ref{tab:pcdtraining}.

\begin{table*}
\centering
\begin{tabularx}{\textwidth}{@{}lc*{7}{>{\centering\arraybackslash}X}@{}}
\toprule
  Total pcd  & Sample Arm & Dropout  & Jitter  & Jitter  & Sample Object  & Pcd  & Pcd  & Grid \\
   points & Points (\#) &  ratio &  ratio &  noise &  Meshes Points &  Normalization &  Scale &  Size\\
  
\midrule
\addlinespace
 6000 & 3000 & [0.1,0.3] & 0.3 & $\mathcal{N}(0,0.01)$ & 1000 & [0,0,0] (toaster) [0.35,0,0.4] (others) & 0.625 (toaster) 1 (others) & 32x32x32\\

\bottomrule
\end{tabularx}
\caption{Point cloud generation and randomization parameters. }
\label{tab:pcddetails}
\end{table*}

\begin{table*}
\centering
\begin{tabularx}{\textwidth}{@{}lc*{6}{>{\centering\arraybackslash}X}@{}}
\toprule
  MLP layers & lr & Optimizer & Batch Size & Nb full pcd traj & Nb simulated pcd traj & Nb simulated pcd traj (distractors) & Nb real traj\\
\midrule
\addlinespace
 256,256 & 0.0003 & AdamW & 32-64 & 15000 & 5000 & 1000 & 15\\

\bottomrule
\end{tabularx}
\caption{ Point cloud teacher-student distillation parameters.  }
\label{tab:pcdtraining}
\end{table*}

\subsection{Simulated Assets Baseline Details}
\label{apdx:simassetsbaseline}
To implement the baseline with multiple simulation assets we had to incorporate two modifications for enabling the multi-task policy learning: 1) at each episode we select a drawer randomly from the set of drawers 2) we expand the observation space of the state-based policy to include the index of the drawer selected to open.

\subsection{Imitation Learning Baseline}

For the imitation learning baseline, we collect 15 (unless otherwise specified) real-world demonstrations using a keyboard interface. We preprocess the point clouds in the same manner as for the teacher-student distillation training (see Section \ref{appdx:teachestudentdistillation}. We complete the point cloud sampling points from the arm mesh leveraging the joints from the real robot. We also add the same randomization: jitter, dropout, and translation. 

\begin{table}
\centering
\begin{tabularx}{\linewidth}{@{}lc*{5}{>{\centering\arraybackslash}X}@{}}
\toprule
 & Pose & Distractors & Disturbances \\ 
 & randomization &  &  \\ 
\midrule
\addlinespace
IL & \textbf{40} $\pm$ 15\% & \textbf{50} $\pm$ 17\% & \textbf{10} $\pm$ 9\%\\ 
IL with distractors &  \textbf{50} $\pm$ 17\% & 20 $\pm$ 13\%  & \textbf{10} $\pm$ 9\% \\ 
\bottomrule
\end{tabularx}
\caption{Comparison of the plain imitation learning baseline (IL) against adding new distractors (IL with distractors) on the task of opening the drawer. No improvement is observed.}
\label{tab:il++}
\vspace{-15pt}
\end{table}

\subsubsection{Imitation learning with new assets}
\label{apdx:il++}
We implemented an additional baseline where we added point clouds sampled from different object meshes (see Figure \ref{fig:distractors}) into the real-world point cloud to make the policy more robust to distractors. However, no improvement in the robustness of this baseline was found as seen in Figure \ref{tab:il++}. We hypothesize that this is the case because the added meshes into the point cloud do not bring any occlusions which is one of the main challenges when adding distractors in point clouds.

\section{Hardware setup}
\label{apdx:hardware}
\begin{figure}[h]
    \centering
    \includegraphics[width=\linewidth]{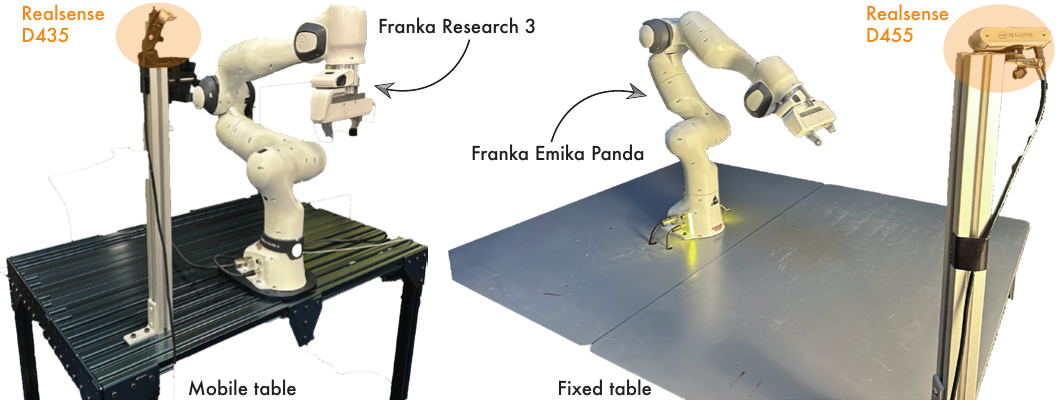}
    \caption{\footnotesize{Overview of the hardware setup used for evaluating \MethodName. left: used for the kitchen toaster task, right: used for the book on the shelf, mug on the shelf, dish on the rack, open cabinet, and open drawer tasks.}}
    \label{fig:robotsetup}
\end{figure}
Our experiments are run on two different Panda Franka arms. One is, the Panda Franka arm 2, which is mounted on a fixed table, we run the book on the shelf, mug on the shelf, dish on the rack, open the cabinet, and open the drawer there. Then we also ran part of our experiments, on a Panda Franka arm 3, mounted on a mobile table, more concretely, the open toaster in the kitchen was the task run on this arm. The communication between the higher and lower level controller of the arm is done through Polymetis \cite{Polymetis2021}. 

We mount one calibrated camera per setup to extract the depth maps that will be passed to our vision policies. More concretely we use the Intel depth Realsense camera D455 on the first setup and the Intel depth Realsense camera D435 on the second setup. See Figure \ref{fig:robotsetup} for more details on the robot setup. 

\section{Further Analysis}
\label{apdx:furtheranalysis}
\subsection{RL from vision}
\label{apdx:rlvision}

Part of the inefficiency of running RL from vision comes from the increased memory required to compute the policy loss for vision-based RL -- on the same GPU, the batch size for vision-based policies is 100x smaller than the batch size used for compact state policies. Rendering point clouds in simulation is also approximately 10x slower than running the pipeline without any rendering. When adding these factors up, RL from vision becomes much slower and practically infeasible given our setup with sparse rewards.
\begin{figure}[h]
    \centering
    \includegraphics[width=\linewidth]{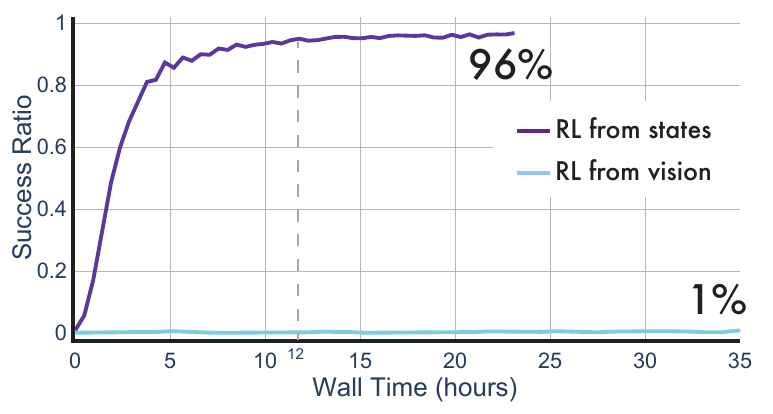}
    \caption{\footnotesize{Wall clock time comparison of running PPO from vision against from compact states.}} 
    \label{fig:visionrlbaseline}
\end{figure}

\subsection{RL from Scratch}

\begin{figure}[h]
    \centering
    \includegraphics[width=\linewidth]{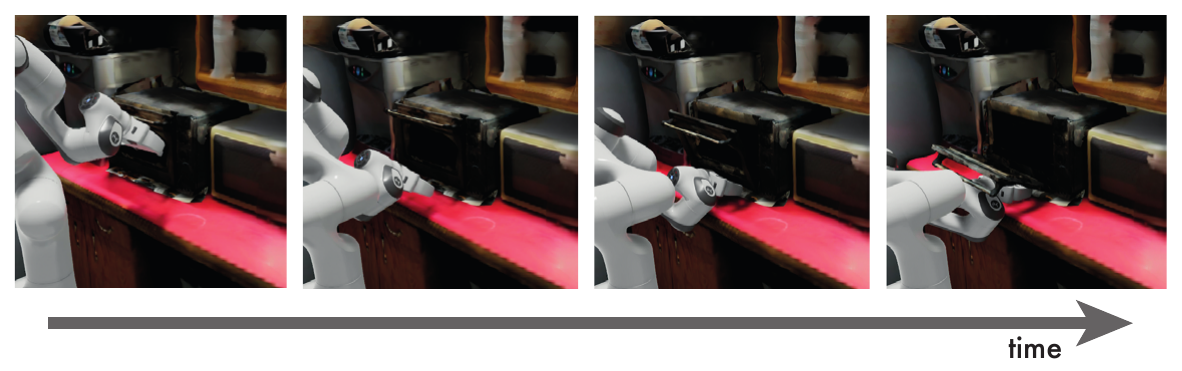}
    \caption{\footnotesize{Visualization of a rollout of the final policy learned with RL without demos and achieving a 62\% accuracy on opening the toaster in simulation. We observe the resulting policy that learns without demos exploits the model's inaccuracies, thereafter it will not transfer to the real world.}}
    \label{fig:rlscratchapdx}
\end{figure}

In Figure \ref{fig:rlscratchapdx}, we qualitatively observe the phenomena that we mention in \ref{sec:rlfinetune}, where the policy trained from scratch, without demos, exploits the model's inaccuracies. In this specific case, we observe that the policy leverages the slightly incorrectly placed joint to open the microwave in an unnatural way that wouldn't transfer to the real world. 

\subsection{RL from different amounts of real-world data}
\label{apdx:rlrealworldablation}
In this section, we analyze further how many real-world demonstrations are needed to successfully fine-tune policies with RL in simulation. We start with 0,5,10,15 real-world demonstrations and inverse-distill the policy by collecting 15 sim trajectories from this real-world trained policy. We observe in table \ref{tab:realworlddemosablation} that for the task of placing a book on the shelf, there is a step function where the PPO has a 0\% success rate until 15 demos are used. The reason is that with less than 15 demos the real-world policy does not transfer to the simulation hence no sim demos can be collected during the inverse distillation procedure. Thereafter the RL fine-tuned policy starts from scratch when using $<$ 15 real-world demos. On the other side, for the easier task of opening a drawer, we observe this step function earlier, where at $>$ 5 demos we can do RL fine-tuning from demos and obtain successful policies. 
\begin{table}
\centering
\begin{tabularx}{\linewidth}{@{}lc*{5}{>{\centering\arraybackslash}X}@{}}
\toprule
 & Book on & Open \\ 
 & shelf & drawer  \\ 
\midrule
\addlinespace
RL fine-tuning from 0 real demos & 0 $\pm$ 0\% & 0 $\pm$ 0\% \\ 
RL fine-tuning from 5 real demos &  0 $\pm$ 0\% & 89 $\pm$ 1\% \\ 
RL fine-tuning from 10 real demos &  0 $\pm$ 0\% & \textbf{96} $\pm$ 1\% \\ 
RL fine-tuning from 15 real demos &  \textbf{90} $\pm$ 2\% & \textbf{96} $\pm$ 1\% \\ 
\bottomrule
\end{tabularx}
\caption{Comparison of training RL from different amounts of real-world demos.}
\label{tab:realworlddemosablation}
\vspace{-15pt}
\end{table}

\subsection{Mixing \MethodName with synthetic data}
\label{apdx:mixsynthtarget}
We run \MethodName combining the data from the synthetic assets experiment (see Figure \ref{fig:simbaseline}) together with the simulated target environment data and study whether we get any performance gain by combining these two sources of data on the task of opening the drawer. We observe in Table \ref{tab:combinesynthetictarget} that there is no clear improvement when combining the simulated assets with the target asset. One reason could be that more synthetic data is needed to observe an increase in performance. The other hypothesis is that learning only on the target environment (\MethodName) is enough and the 10\% left to reach 100\% success rate in the real world comes from the sim-to-real gap.

\begin{table}
\centering
\begin{tabularx}{\linewidth}{@{}lc*{5}{>{\centering\arraybackslash}X}@{}}
\toprule
 & Pose & Distractors  \\ 
 & randomization & \\ 
\midrule
\addlinespace
\MethodName & \textbf{90} $\pm$ 9\% & \textbf{90} $\pm$ 9\%\\ 
\MethodName + synthetic assets &  \textbf{90} $\pm$ 9\% & \textbf{80} $\pm$ 13\% \\ 
\bottomrule
\end{tabularx}
\caption{Comparison of using \MethodName with added synthetic assets against standard \MethodName on the task of opening the drawer in the real world. No improvement is observed.}
\label{tab:combinesynthetictarget}
\vspace{-15pt}
\end{table}

\subsection{\MethodName Multi-Task}
\label{apdx:rialtomultitask}

We propose a \textit{multi-task} version of \MethodName. We train \textit{multi-task} \MethodName on the tasks of opening a drawer, putting a mug on the shelf, cup in the trash, and dish on the rack environments.

The proposed \textit{multi-task} \MethodName procedure is the following: 
\begin{enumerate}
    \item  Train separate state-based single-task policies per task
    \item  Collect trajectories from each one of the tasks with the state-based policies
    \item  Distill these trajectories into a single multi-task policy conditioned with the task-id
    \item Run multiple iterations of DAgger on each task sequentially to obtain a final multi-task policy
\end{enumerate}

We evaluate this policy in the real world on two of the tasks and observe in Table \ref{tab:multitaskrialto} that in \emph{opening the drawer}, the performance of multi-task \textbf{RialTo} matches single-task (90\% success). However, the performance slightly decreases on the \emph{mug on the shelf task} (from 100\% on single-task to 80\% on multi-task). Nevertheless, the performance is still above the imitation learning baseline (40\% for the drawer and 10\% for the mug on the shelf).
We did not tune any hyperparameters, and we kept the same network size that we used for the \textbf{RialTo} experiments. We should be able to bring the performance of the mug on the shelf task to match the single-task policy with some hyperparameter tuning. 
\\
We showed that \textbf{RialTo} can be easily adapted to train multi-task policies. We hypothesize that we need to train in more environments to obtain multi-task generalization.
\begin{table}
\centering
\begin{tabularx}{\linewidth}{@{}lc*{5}{>{\centering\arraybackslash}X}@{}}
\toprule
 & Open & Mug  \\ 
 & drawer& on shelf \\ 
\midrule
\addlinespace
Imitation learning  & 40 $\pm$ 17\% & 10 $\pm$ 9\%\\ 
\MethodName & \textbf{90} $\pm$ 9\% & \textbf{100} $\pm$ 0\%\\ 
\MethodName multitask &  \textbf{90} $\pm$ 9\% & 80 $\pm$ 15\% \\ 
\bottomrule
\end{tabularx}
\caption{Comparison of training \MethodName on multiple tasks against single-task \MethodName. No improvement is observed.}
\label{tab:multitaskrialto}
\vspace{-15pt}
\end{table}

\subsection{Sim-to-real gap}
\label{apdx:sim2realgap}

We analyze and propose an explanation for the observed sim-to-real gap in Table \ref{tab:sim2realgap}, where we show the performance of the final point cloud-based policy in both simulation and the real world.
We observe that in general, the sim-to-real gap does not seem to be present.
In some cases such as for the mug on shelf task, we observe that the performance in simulation is worse than the performance in the real world. The main reason for this disparity is that we want to make the simulation harder than the real-world environment to make sure that we will be able to recover a good robust policy in the real world.

\begin{table*}[ht]
\centering
\begin{tabularx}{\textwidth}{@{}lc*{6}{>{\centering\arraybackslash}X}@{}}
\toprule
 & Kitchen & Book on & Plate on & Mug on & Open & Open \\
 & toaster & shelf & rack & shelf & drawer  &  cabinet  \\
\midrule
\addlinespace
Performance in simulation & 90 $\pm$ 4\% & 84 $\pm$ 5\% & 80 $\pm$ 6\% & 72 $\pm$ 6\% & 95 $\pm$ 3\% & 92 $\pm$ 4\% \\
Performance in the real world & 90 $\pm$ 9\% & 90 $\pm$ 9\% & 90 $\pm$ 9\% & 100 $\pm$ 0\% & 90 $\pm$ 9\% & 85 $\pm$ 8\% \\

\bottomrule
\end{tabularx}
\caption{Comparison of performance in simulation (top) and the real world (bottom).}
\label{tab:sim2realgap}
\end{table*}

\section{GUI for Real-to-Sim Transfer of Scenes}
\label{apdx:r2simscenes}
In the main text and video, we provide an overview of the features and capabilities of our GUI. Additional valuable features include the ability to populate the scene with assets from object datasets such as Objaverse~\cite{deitke2023objaverse}. This allows for randomizing surrounding clutter and supporting policy training that generalizes to distractor objects (see Section~\ref{subsec:rldistractors}).

\subsubsection{3D reconstruction software used} We mainly used 3 different methods/apps for obtaining the 3D meshes from videos: 
\begin{enumerate}
    \item Polycam \cite{polycam2020} is used to scan larger scenes, such as the kitchen. Polycam makes effective use of the built-in iPhone depth sensor which helps extract realistic surface geometry for large uniform flat surface (e.g., a kitchen counter). However, we find it struggles with fine-grained details. Polycam outputs a GLTF file, which we convert directly to a USD for loading into Isaac Sim using an online conversion tool.
    \item  AR Code \cite{arcode2022} is used to extract high-quality meshes for single objects that can be viewed by images covering the full 360 degrees surrounding the object (e.g., cabinet, mug, microwave, drawer). While AR Code leads to more accurate geometry than Polycam for singulated objects, we still find it struggles on objects with very thin parts. AR Code directly outputs a USD file that can be loaded into Isaac Sim. 
    \item NeRFStudio  \cite{tancik2023nerfstudio} is used to capture objects that require significantly more detail to represent the geometry faithfully. For example, AR Code failed to capture the thin metal structures on the dish rack, whereas NeRFs are capable of representing these challenging geometric parts. We use the default ``nerfacto'' model and training parameters. This method trains a relatively small model on a single desktop GPU in about 10 minutes. After training converges, we use the NeRFStudio tools for extracting a 3D point cloud and obtaining a textured mesh with Poisson Surface Reconstruction~\cite{kazhdan2006poisson}. This outputs an OBJ file, which we convert into a USD by first converting from OBJ to GLTF, and then converting from GLTF into USD (with both file conversions performed with an online conversion tool).
\end{enumerate}

\section{GUI User study}
\label{apdx:userstudy}

\begin{table*}[ht]
\centering
\begin{tabularx}{\textwidth}{@{}lc*{8}{>{\centering\arraybackslash}X}@{}}
\toprule
 & Scan & Process + Upload 1st Scan (idle)  & Cut & Joint & 2nd Scan & Process + Upload 2nd Scan (idle) & Total time & Total active time\\
\midrule
\addlinespace
User 1 & 2:25 & 5:41 & 4:15 & 4:56 & 8:10 & 10:45 & 36:12 & 19:46 \\
User 2 & 6:30 & 12:57 & 3:32 & 3:51 & 2:37 & 4:19 & 33:46 & 16:30 \\
User 3 & 3:52 & 5:52 & 4:35 & 4:14 & 3:26 & 4:15 & 26:14 & 16:07 \\
User 4 & 2:34 & 2:06 & 2:48 & 1:41 & 5:14 & 4:33 & 19:06 & 12:27 \\
User 5 & 1:32 & 2:33 & 4:43 & 1:28 & 4:34 & 3:50 & 18:40 & 12:17 \\
User 6 & 2:30 & 3:52 & 2:08 & 1:17 & 4:59 & 2:26 & 17:12 & 10:54\\

\bottomrule
\end{tabularx}
\caption{Detailed time spent by each user in the user study, see Section \ref{sec:userstudies}.}
\label{tab:userstudy}
\end{table*}

To test the functionality and versatility of the real-to-sim generation pipeline, we ran a user study over six people, where each participant was tasked with creating an articulated scene using the provided GUI. Every individual was given the same set of instructions that would guide them through the process of constructing a usable and accurate scene. At the start of each trial, the participant was instructed to download Polycam \cite{polycam2020}, which uses a mobile device’s LiDAR to generate 3D models. The user then selected a location and captured their scene by taking a sequence of images. The time required to complete this step was recorded as “Scan Time.” Once the images were captured, Polycam needed to process the pictures to transform the scene into a three-dimensional mesh. Once the mesh had been generated, the participant was then instructed to upload the articulated USD to a computer and convert this file into the GLB format (required by our GUI). Finally, the user uploaded the GLB file into the provided GUI, and the time required to complete these steps was recorded as “Scan Processing and Uploading Time.” Because the uploaded mesh was created using one scan, all objects in the scene are connected, and the user is unable to move a single item without shifting the entire background. Thus, in order to create a more realistic scene, the participant was asked to use the GUI to cut an object out of the scene, allowing this item to be manipulated independently of the background. The time it took for the user to cut this object from the original mesh was regarded as “Cut Time.” In an attempt to further the realistic nature of this scene, the participant was then instructed to specify joint parameters and create a fixed joint that would allow an object in the scene to rotate about a specific point. For instance, a fixed joint at a door would allow the door to rotate about its hinge and generate an accurate simulation of door movement. The time required to create a fixed joint in the scene was recorded as “Joint Time.” Lastly, to demonstrate the full capabilities of the GUI, the participant was asked to add another object to their current scene. They were instructed to download another 3D scanning application, AR Code \cite{arcode2022}, which was used to create the three-dimensional mesh of the additional object. The time required to generate this mesh was recorded as “Scan Time (2).” Then the participant again converted their mesh to GLB format and uploaded this file to the same GUI. Once uploaded, the object was placed in a realistic position within the scene, and the time elapsed during this step was added to the “Scan Processing and Uploading Time” category. Through this user study, we found that it took an average of 14.67 active minutes (excluding the “Scan Processing and Uploading Time” category) to create a scene that included one cut object, one fixed joint, and one additional object. However, it is important to note that User 6 had previous experience using this GUI, while all other users had no experience. Thus, if we disregard the results of User 6, we find the average time to create a scene to be 15.42 active minutes, which is not a significant difference. As a result, the real-to-sim transfer using the provided GUI seems to be an intuitive process that is neither time nor labor-intensive. 

\textbf{User 1} took the longest time to complete this series of tasks mostly due to their extensive upload period. Because User 1 scanned their environment for a lengthy period, their articulated USD file was larger than all other users. As a result, it took longer for them to upload their file to a computer and convert this file to GLB format. The abnormal size of User 1’s file coupled with their difficulty operating the file conversion website led to a lengthy Scan Processing and Upload Time, which led to the slowest overall  performance.

\textbf{User 2} was the only user who was sent instructions digitally and completed the tasks remotely. An individual experienced with the real-to-sim pipeline was present for all other trials. Thus, this may have contributed  to User 2’s longer completion time, as their questions had to be answered remotely. However, User 2 did not have trouble with any particular section of the pipeline but rather took a longer time to complete each section.

\textbf{User 3}’s experience with the real-to-sim pipeline went smoothly, as there were no obvious difficulties while scanning, uploading, or using the GUI. They followed the instructions quickly and precisely, resulting in a better completion time than Users 1 and 2.

\textbf{Users 4 and 5} completed all tasks in the pipeline more quickly than User 3 because the background they chose was smaller with fewer details. Thus, they were able to scan their scenes faster, generating a smaller file that was able to be processed, uploaded, and converted more quickly. However, their speed did reduce the quality of their backgrounds, since the details in both scans are not as precise as the others. Thus, it seems User 3 completed the tasks quickly with the most accurate scan.

\textbf{User 6} had previous experience with the real-to-sim pipeline, so they were able to use this expertise to quickly complete the tasks. The only abnormality with User 6’s trial was their longer Scan Time for object 2. They had trouble with the “AR code” app during this trial, resulting in a longer Scan Time (2).

\subsection{Scaling laws of the RialTo GUI}
\label{apdx:scalinglawsscanning}
\begin{equation}
    \label{eq:scaling}
    \begin{split}
                total\_active\_time = t_{\textit{scan scene}} \\+ t_{\textit{scan object}} \cdot N_{\textit{objects}}\\ + t_{\textit{cut object}} \cdot N_{\textit{cut objects}}\\ + t_{\textit{add joint}} \cdot N_{joints}
    \end{split}
\end{equation}

We derive a relation to express the total active time needed to create a scene with respect to the number of joints and objects there are in the scene. The total active time to create a scene increases linearly in complexity with the number of objects and joints present in the scene, as seen in Relation \ref{eq:scaling}. We define $N_\textit{objects}$ as the number of scanned objects that we want to add, $N_\textit{cut objects}$ as the number of objects that we want to extract from the scanned scene, $N_{\textit{joints}}$ as the number of joints the scene has. Taking the average times from our user study (see Table \ref{tab:userstudy}) we find $t_{\textit{scan object}} = 4:50$, $t_{\textit{scan scene}} = 3:14$, $t_{\textit{add joint}} = 2:54$, $t_{\textit{cut object}} = 3:40$. Note that these values are on the conservative side since only one user was an expert, and with increased expertise, these coefficients become smaller.

\begin{equation*}
\end{equation*}

\section{Compute resources}
\label{apdx:compute}
We run all of our experiments on an NVIDIA GeForce RTX 2080 or an NVIDIA GeForce RTX 3090. The first step of learning a vision policy from the real-world demos and collecting a set of 15 demonstrations in simulation takes an average of 7 hours. The next step of RL fine-tuning from demonstrations takes on average 20 hours to converge. Finally, the teacher-student distillation step takes 24 hours between collecting the trajectories, distilling into the vision policy, and running the last step of DAgger. This adds up to a total of 2 days and 3 hours on average to train a policy for a given task.

\end{document}